%% file: main.tex
\documentclass{article} 
\usepackage{longcat_style,times}

\input{math_commands.tex}

\usepackage{hyperref}
\usepackage{url}

\usepackage{graphicx}
\usepackage{balance}  
\usepackage{amsmath}
\usepackage{algpseudocode}
\usepackage{latexsym}
\usepackage{multirow}
\usepackage{multicol}
\usepackage{color}
\usepackage{dashrule}
\usepackage{extarrows}
\usepackage{dsfont}
\usepackage{centernot}
\usepackage{hyperref}
\usepackage{natbib}
\usepackage{fancybox}
\usepackage[utf8]{inputenc} 
\usepackage[T1]{fontenc}    
\usepackage{hyperref}       
\usepackage{url}            
\usepackage{booktabs}       
\usepackage{amsfonts}       
\usepackage{amsmath}
\usepackage{nicefrac}       
\usepackage{microtype}      
\usepackage{xcolor} 
\usepackage{amsmath}
\usepackage{lipsum}
\usepackage{colortbl}
\usepackage{subcaption}
\usepackage{tabularx}
\usepackage{makecell}
\usepackage{wrapfig}
\usepackage{bm}
\usepackage{multirow}
\usepackage{xcolor} 
\usepackage{bbm}
\usepackage{mdframed}
\usepackage{enumitem}
\usepackage{xspace}

\usepackage{lineno}

\usepackage{xurl}
\usepackage{subcaption}
\usepackage{amsmath}
\usepackage{multirow}
\usepackage{makecell}
\usepackage{booktabs}
\usepackage{bm}

\usepackage[T1]{fontenc}

\usepackage[utf8]{inputenc}

\usepackage{microtype}

\usepackage{inconsolata}

\usepackage{graphicx}

\usepackage{amsmath}
\usepackage{amssymb}
\usepackage{booktabs}
\usepackage{multirow}
\usepackage{makecell}
\usepackage{tcolorbox}
\usepackage{listings}
\usepackage{xcolor}
\usepackage[table]{xcolor} 
\usepackage{graphicx}
\usepackage{amsmath}

\definecolor{promptbg}{RGB}{248,248,248}
\definecolor{promptframe}{RGB}{210,210,210}

\lstdefinestyle{promptstyle}{
  basicstyle=\ttfamily\small,
  columns=fullflexible,
  breaklines=true,
  breakatwhitespace=true,
  keepspaces=true,
  showstringspaces=false,
  frame=none,
  tabsize=2,
  gobble=0,
  breakindent=0pt,         
  breakautoindent=false,   
  postbreak=\mbox{}        
}

\newtcolorbox{promptbox}[1][Prompt]{
  colback=promptbg,
  colframe=promptframe,
  title=\textbf{#1},
  boxrule=0.6pt,
  arc=1mm,
  coltitle=black,        
  colbacktitle=gray!25,  
  boxsep=4pt,            
  left=4pt, right=4pt, top=4pt, bottom=4pt 
}
\definecolor{darkblue}{rgb}{0, 0, 0.5}
\hypersetup{colorlinks=true, citecolor=darkblue, linkcolor=darkblue, urlcolor=darkblue}

\usepackage{listings}
\usepackage{xcolor}
\usepackage[ruled,vlined]{algorithm2e}
\usepackage{caption}
\usepackage{float} 
\usepackage{arydshln}
\usepackage{wrapfig} 
\usepackage{tcolorbox}
\tcbuselibrary{theorems}
\tcbuselibrary{most}
\usepackage[dvipsnames]{xcolor}

\usepackage{subcaption}
\usepackage{multirow}
\usepackage{tcolorbox}
\usepackage{xcolor}
\usepackage{csquotes}
\usepackage{pifont}

\usepackage[table]{xcolor}
\tcbuselibrary{skins, breakable}

\definecolor{customTeal}{RGB}{0, 128, 128} 
\definecolor{emphasisColor}{RGB}{255, 0, 0} 
\definecolor{oursBlue}{RGB}{51,202,246}
\newcommand{\name}{\textsc{Attention-MoA}\xspace}
\definecolor{blue1}{HTML}{508AB2}
\definecolor{green2}{HTML}{BFF6BA}

\newtcbtheorem[]{prompt}{Prompt}%
{colback=SeaGreen!10!CornflowerBlue!10,
 colframe=RoyalPurple!55!Aquamarine!100!,
 fonttitle=\bfseries,
 left=.02in, right=.02in, bottom=.02in, top=.02in,
 before upper={\linespread{1.5}\selectfont}}{prompt}

\renewcommand{\headeright}{\raisebox{-0.2\height}{\includegraphics[height=2.2em]{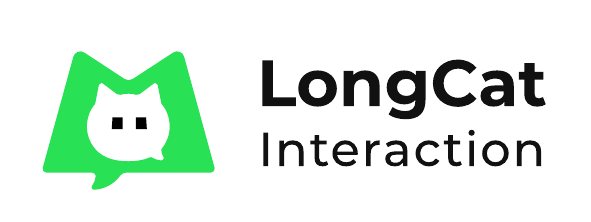}}}
\renewcommand{\shorttitle}{Attention-MoA}

\definecolor{darkblue}{rgb}{0, 0, 0.5}
\hypersetup{colorlinks=true, citecolor=darkblue, linkcolor=darkblue, urlcolor=darkblue}

\makeatletter
\renewcommand{\@maketitle}{%
  \vbox{%
    \hsize\textwidth
    \linewidth\hsize
    \vskip -0.5in
    \noindent
    \begin{minipage}{0.2\textwidth}
      \includegraphics[width=\linewidth]{pic/Logo.pdf}
    \end{minipage}%
    \\
    \rule{\linewidth}{1pt}
    \hspace{0.05\textwidth}%
    \begin{minipage}{0.8\textwidth}
    \end{minipage}

    \centering
    {\LARGE \bfseries\@title\par}
    \vskip 0.1in  
    \def\And{%
      \end{tabular}\hfil\linebreak[0]\hfil%
      \begin{tabular}[t]{c}\bf\rule{\z@}{24\p@}\ignorespaces%
    }
    \def\AND{%
      \end{tabular}\hfil\linebreak[4]\hfil%
      \begin{tabular}[t]{c}\bf\rule{\z@}{24\p@}\ignorespaces%
    }
    \begin{tabular}[t]{c}\bf\rule{\z@}{24\p@}\@author\end{tabular}%
  \vskip 0.05in 
  }
}
\makeatother

\title{\name: Enhancing Mixture-of-Agents via Inter-Agent Semantic Attention and Deep Residual Synthesis \\}

\makeatletter
\def\@fnsymbol#1{\ensuremath{\ifcase#1\or \dagger\or \ddagger\or
   \mathsection\or \mathparagraph\or \|\or **\or \dagger\dagger
   \or \ddagger\ddagger \else\@ctrerr\fi}}
\makeatother

\author{
\begin{tabular}{c}
\textbf{Jianyu Wen}
\quad
\textbf{Yang Wei}$^\dagger$
\quad
\textbf{Xiongxi Yu}
\quad
\textbf{Changxuan Xiao}
\quad
\textbf{Ke Zeng}\thanks{Ke Zeng and Yang Wei are the corresponding authors.} \\[1ex]
\quad
Meituan LongCat Interaction Team\\[1ex]
\quad
\normalfont \texttt{\{wenjianyu, weiyang14, yuxiongxi\}@meituan.com}\\[1ex]
\quad
\normalfont \texttt{\url{https://github.com/John-Wendell/Attention-MoA}}
\end{tabular}
}

\begin{document}

\maketitle

\input{sections/abstract.tex}

\input{sections/introduction.tex}
\input{sections/methodology.tex}
\input{sections/experiment.tex}

\input{sections/conclusion}

\bibliographystyle{iclr2026_conference}
\bibliography{ijcai26}

\newpage
\appendix

\input{sections/appendix}

\end{document}

%% file: math_commands.tex
\usepackage{amsmath,amsfonts,bm}

\def\eqref#1{equation~\ref{#1}}

\def\1{\bm{1}}

\DeclareMathAlphabet{\mathsfit}{\encodingdefault}{\sfdefault}{m}{sl}
\SetMathAlphabet{\mathsfit}{bold}{\encodingdefault}{\sfdefault}{bx}{n}



%% file: sections/abstract.tex
\begin{abstract}
As the development of Large Language Models (LLMs) shifts from parameter scaling to inference-time collaboration, the \textbf{Mixture-of-Agents (MoA)} framework has emerged as a general paradigm to harness collective intelligence by layering diverse models.
While recent MoA variants have introduced dynamic routing and residual connections to improve efficiency, these methods often fail to facilitate deep semantic interaction between agents, limiting the system's ability to actively correct hallucinations and refine logic. 
In this paper, we introduce \textbf{Attention-MoA}, a novel MoA-based framework that redefines collaboration through \textbf{Inter-agent Semantic Attention}. 
Complemented by an \textbf{Inter-layer Residual Module} with \textbf{Adaptive Early Stopping Mechanism}, our architecture mitigates information degradation in deep layers while improving computational efficiency.
Extensive evaluations across AlpacaEval 2.0, MT-Bench, and FLASK demonstrate that Attention-MoA significantly outperforms state-of-the-art baselines, achieving a 91.15\% Length-Controlled Win Rate on AlpacaEval 2.0 and dominating in 10 out of 12 capabilities on FLASK.
Notably, Attention-MoA enables an ensemble of small open-source models to outperform massive proprietary models like Claude-4.5-Sonnet and GPT-4.1, achieving an MT-Bench score of 8.83 and an AlpacaEval 2.0 LC Win Rate of 77.36\%.
\end{abstract}

%% file: sections/introduction.tex
\vspace{-10pt}
\section{Introduction}

The advancement of LLMs has revolutionized artificial intelligence, with large-scale models like the GPT, Claude, and Gemini series demonstrating remarkable capabilities. As parameter counts approach the trillion scale, the focus is moving from scaling individual model parameters to optimizing inference-time collaboration~\cite{Guo2024multiagentsurveyijcai,ji2025survey,tran2025multi,Wei2025ijcaiContextAware,liu2025survey}. 
For models of this magnitude, further optimizing the training phase faces prohibitive computational costs and diminishing returns. 
Consequently, multi-agent collaboration has emerged as a vital strategy to overcome the inherent limitations of monolithic models—such as hallucinations—by leveraging collective intelligence. 
To situate our contributions within the current landscape, we first review the evolution of these collaborative architectures, progressing from unstructured ensembles to structured layered frameworks.

The initial exploration of this paradigm focused on \textbf{Multi-Agent Ensembles (MAE)}. This approach operates on the \textit{scaling law of agents}, where simple aggregation mechanisms allow groups to outperform their strongest individual members. 
Research demonstrates that ensemble methods can unlock reasoning capabilities inaccessible to single models by mitigating individual variance and hallucinations through techniques like majority voting \cite{li2024more,yang2025multillmcollaborativesearchcomplex,chen2024more} or aggregation model~\cite{tang2025open,chen2025symbolic,li2025llms}.

To address the lack of interaction in ensembles, \textbf{Multi-Agent Debate (MAD)} was introduced to emphasize iterative critique and consensus. 
MAD established that allowing models to critique and correct one another significantly reduces hallucinations \cite{liyun2024improving,liang-etal-2024-encouraging,Akbir2024Debating,xu2025generative}. Protocols like round-table conferences force diverse LLMs toward a reasoned consensus rather than a simple aggregate \cite{chen-etal-2024-reconcile}. 
DEEVO \cite{nair2025tournament} utilizes structured debates and Elo ratings as a fitness proxy to evolve high-quality prompts without ground truth.
However, debate frameworks face the risk of conformity, where highly persuasive but incorrect models mislead the group \cite{Akbir2024Debating,nair2025tournament}, or infinite loops of disagreement.

Most recently, the \textbf{Mixture-of-Agents (MoA)} framework~\cite{wang2024mixtureofagentsenhanceslargelanguage} introduced a multi-layered architecture to MAE, expanding the collaborative paradigm beyond the closed-form tasks with definitive answers or finite solution spaces of MAD (e.g., fact-checking).
The inherent generalizability for both open-form and closed-form tasks of MoA has facilitated its widespread application, encompassing clinical informatics\cite{gao2025moma,jang2025yalenlp}, software engineering\cite{sharma2024patched,ashiga2025industrial,ping2025verimoa}, and the broader industrial landscape\cite{mitra2024distributed,chen2025tumix,chakraborty2025collab}.
While the standard MoA framework introduced a promising collaborative paradigm, its naive concatenation mechanism and fixed architecture have necessitated a series of targeted optimizations.
To mitigate computational inefficiency, Sparse-MoA~\cite{li2025smoa} introduced a gating mechanism and early stopping. 
Sparse-MoA significantly reduced token consumption, albeit with the overhead of maintaining additional control agents.
Furthermore, RMoA~\cite{xie-etal-2025-rmoa} integrated residual connections and diversity selection based on embedding. 
Inspired by ResNet, it employs residual agents to minimize information loss and introduces an adaptive termination mechanism to improve efficiency.
Self-MoA~\cite{li2025rethinking} challenged the necessity of mixing heterogeneous models. 
It demonstrated that naive mixing often degrades performance due to inconsistent agent quality. 
Instead, it proposed aggregating multiple samples from a single high-quality model to maximize coherence.
Despite these advancements, the MoA paradigm generally lacks inter-agent communication mechanisms and effective interaction between outputs across different rounds. 

In this paper, we introduce \textbf{Attention-MoA}, a novel MoA-based framework designed to address these limitations through Inter-agent Semantic Attention and Inter-layer Residual Connections. 
Unlike standard MoA which relies on naive concatenation, our Attention Module facilitates a semantic critique-and-refine process. 
Through cross-agent and self-attention, agents generate explicit natural language instructions to critique and refine peer responses, ensuring that the intra-layer summary is synthesized from locally optimized reasoning. 
Simultaneously, the Residual Module maintains a historical context stack to prevent information degradation across layers, while incorporating an adaptive early stopping mechanism to optimize inference efficiency and reduce computational costs by dynamically terminating redundant reasoning cycles.
Our contributions are summarized as follows:

\begin{itemize}

    \item \textbf{Attention-MoA Architecture:} We propose a novel collaborative framework that synergies Inter-agent Semantic Attention with an Inter-layer Residual Module. The former utilizes natural language instructions to facilitate explicit peer-critique and self-refinement, while the latter accumulates historical context to prevent information degradation. This design enables the system to translate architectural depth into scalable performance gains.
    \item \textbf{Adaptive Inference Efficiency:} To optimize computational resource allocation, we introduce an Adaptive Early Stopping mechanism embedded within the residual synthesis process. By dynamically evaluating convergence and trajectory, this mechanism eliminates redundant iterations, reducing the consumption of inference tokens by approximately 11\%.
    \item \textbf{State-of-the-Art Performance:} Extensive evaluations demonstrate that Attention-MoA significantly outperforms existing baselines. 
    On AlpacaEval 2.0, our method achieves Length-Controlled Win Rate of 91.15\%, and on MT-Bench, it reaches a score of 9.32. 
    Furthermore, fine-grained analysis on the FLASK reveals that our method achieves superior performance in 10 out of 12 dimensions. 
    Notably, it establishes a dominant lead in critical cognitive and safety metrics—including \textit{Harmlessness}, \textit{Factuality}, \textit{Correctness}, \textit{Comprehension}, \textit{Commonsense}, \textit{Robustness}, \textit{Readability}, \textit{Metacognition} and \textit{Insightfulness}.
    \item \textbf{Generalizability to Small-Scale Models:} We empirically demonstrate that Attention-MoA effectively scales down to smaller, open-source models with active parameters ranging from 12B to 32B. 
    Remarkably, our small-scale configuration achieves MT-Bench score of 8.83 and AlpacaEval 2.0 Win Rate of 77.36\%. These results surpass massive proprietary systems like Claude-4.5-Sonnet (8.62 / 73.49\%) and GPT-4.1 (8.59 / 69.83\%).

\end{itemize}

\begin{figure}[tb]
    \centering
    \begin{subfigure}[b]{0.7\textwidth}
        \centering
        \includegraphics[width=\linewidth]{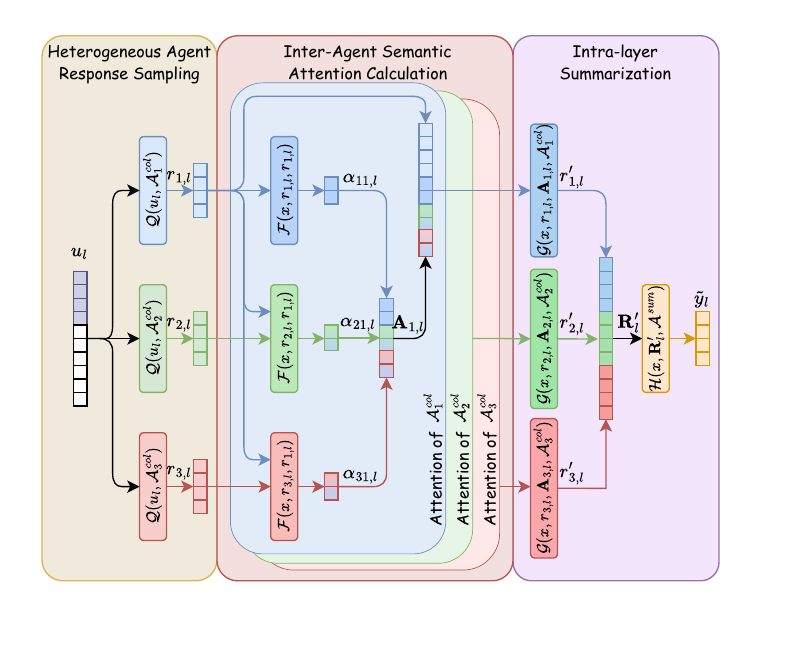}
        \caption{Inter-agent Attention Module (Intra-layer)}
        \label{fig:intra_layer}
    \end{subfigure}
    \begin{subfigure}[b]{0.7\textwidth}
        \centering
        \includegraphics[width=\linewidth]{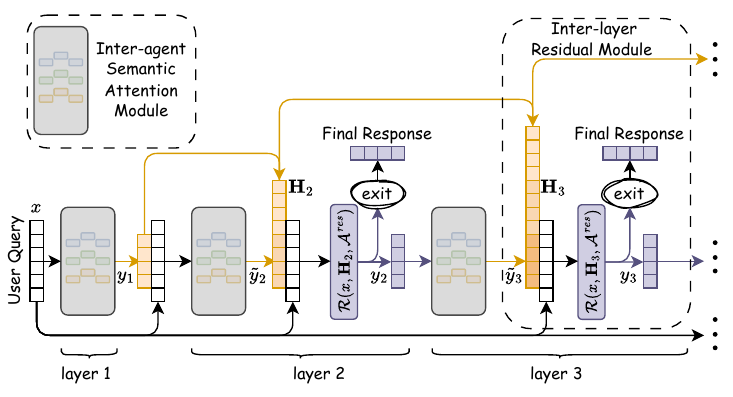}
        \caption{Inter-layer Residual Module (Inter-layer)}
        \label{fig:inter_layer}
    \end{subfigure}
    \caption{\textbf{Overview of the Attention-MoA Framework.} 
    (a) The intra-layer workflow where heterogeneous collaborative agents ($\mathcal{A}^{col}$) generate responses refined by  inter-agent semantic attention mechanism and Intra-layer Summary Agent ($\mathcal{A}^{sum}$). 
    (b) The inter-layer residual pathway where residual agents ($\mathcal{A}^{res}$) incorporate an adaptive early stopping mechanism to dynamically control inference depth.}
    \label{fig:framework_overview}
\end{figure}

%% file: sections/methodology.tex
\vspace{-10pt}
\section{Methodology}

In this section, we present the Attention-MoA framework. 
As shown in Figure~\ref{fig:framework_overview}, the architecture is designed as a multi-layer system consisting of $L$ layers, where the configuration evolves with the network depth to optimize information flow and error correction.
The structure is defined as follows:
\begin{itemize}
    \item \textbf{First Layer ($l=1$):} 
    This layer consists solely of the Inter-agent Semantic Attention Module. It takes the initial user query $x$ as input, and the output of this module serves directly as the output, denoted as $y_1$.
    \item \textbf{Subsequent Layers ($l \ge 2$):} 
    Each of these layers integrates two mechanisms: the Inter-agent Semantic Attention Module followed by the Inter-layer Residual Module. 
    The Attention Module in layer $l$ receives an input consisting of the initial query $x$ and the output from the previous layer $y_{l-1}$. The output of the Attention Module is then passed to the Residual Module, which synthesizes the final output $y_l$ for the current layer.
\end{itemize}

This design ensures that early layers focus on generating diverse perspectives, while deeper layers leverage historical context to refine consensus and prevent information degradation. The implementation details are elaborated below.

\subsection{Inter-agent Semantic Attention Module}

To mitigate the hallucinations and logical fallacies in single model, this module introduces an intra-layer collaboration method (Figure \ref{fig:intra_layer}). 
By simulating critique-and-refine processes, agents dynamically evaluate peers within one layer.

\subsubsection{Heterogeneous Agent Response Sampling}


In the initial phase of collaboration, diversity is paramount. We collect a set of heterogeneous agents, $\mathcal{A}^{col}_1, \dots, \mathcal{A}^{col}_N$, referred to as \textbf{Collaborative Agents}.

For a specific layer $l$, the input $u_l$ provided to the agents depends on the depth of the layer:

$$
u_l = 
\begin{cases} 
x & \text{if } l = 1 \\
\{x, y_{l-1}\} & \text{if } l > 1 
\end{cases}
$$

where $x$ is the initial query and $y_{l-1}$ is the output from the preceding layer. Note that for $l=2$, $y_1$ is the direct output of the first layer's Attention Module; for $l > 2$, $y_{l-1}$ is the output of the previous layer's Residual Module.
Each agent $\mathcal{A}^{col}_i$ generates its initial response $r_{i,l}$ in parallel:

$$ r_{i,l} = \mathcal{Q}(u_l, \mathcal{A}^{col}_i), \quad \forall i \in \{1, \dots, N\} $$

This Heterogeneous Sampling Strategy $\mathcal{Q}$, ensures that the response space encompasses diverse perspectives while maintaining alignment with the original query.

\subsubsection{Inter-Agent Semantic Attention Calculation}

We generalize the traditional attention mechanism from inter-embedding to Inter-Agent Semantic Attention. 
This attention represents the semantic influence one agent exerts on another. 
Specifically, $\mathcal{A}^{col}_i$ formulates a refinement suggestion $\alpha_{ij,l}$ for $r_{j,l}$, leveraging its own response $r_i$ as a reference.

We define the attention calculation based on the relationship between the Advisor ($\mathcal{A}^{col}_i$) and the Recipient ($\mathcal{A}^{col}_j$):

\begin{itemize}
    \item \textbf{Cross-Attention ($i \neq j$):} $\mathcal{A}^{col}_i$ uses its own response $r_{i,l}$ as the \textit{Query} to attend to the peer response $r_{j,l}$ . By comparing the two perspectives, $\mathcal{A}^{col}_i$ generates a corrective instruction $\alpha_{ij,l}$. This instruction highlights discrepancies, identifies hallucinations in $r_{j,l}$, or suggests incorporating valuable insights from $r_{i,l}$ into $r_{j,l}$.
    \vspace{4pt}
    \item \textbf{Self-Attention ($i = j$):} $\mathcal{A}^{col}_i$ critically re-evaluates $r_{i,l}$ to identify internal inconsistencies or potential errors, generating a self-refinement instruction $\alpha_{ii,l}$.
\end{itemize}

Formally, the attention $\alpha_{ij,l}$ is generated by the attention function $\mathcal{F}$:

$$ \alpha_{ij,l} = \mathcal{F}(x, r_{i,l}, r_{j,l}, \mathcal{A}^{col}_i) $$

In contrast to traditional scalar attention weights, $\alpha_{ij,l}$ manifests as a natural language instruction that explicitly guides the refinement of $r_{j,l}$.

Following the generation of these  semantic attention, each agent $\mathcal{A}^{col}_j$ collects the set of all semantic attentions directed at it, denoted as $\mathbf{A}_{j,l} = \{\alpha_{1j,l}, \alpha_{2j,l}, \dots, \alpha_{Nj,l}\}$. 
This set contains both the self-attention insight and the cross-attention critiques from all other peers. The agent then synthesizes these instructions to revise its original response $r_{j,l}$ via the update function $\mathcal{G}$:

$$ r'_{j,l} = \mathcal{G}(x, r_{j,l}, \mathbf{A}_{j,l}, \mathcal{A}^{col}_j) $$

This \textbf{Semantic Attention Aggregation} ensures that every agent's output is optimized, having corrected its own errors and absorbed relevant details from the group consensus.





\subsubsection{Intra-layer Summarization}

To unify the refined responses, the \textbf{Intra-layer Summary Agent}, $\mathcal{A}^{sum}$, acts as the projection layer of traditional attention mechanism. 
It takes the refined responses $\mathbf{R}'_l$ as input and synthesizes the attention module output, denoted as $\tilde{y}_l$:

$$ \tilde{y}_l = \mathcal{H}(x,\mathbf{R}'_l, \mathcal{A}^{sum}) $$

For the first layer, this output of the module becomes the output of the layer (i.e., $y_1 = \tilde{y}_1$). For subsequent layers ($l \ge 2$), $\tilde{y}_l$ serves as the input to the Inter-layer Residual Module.

\subsection{Inter-layer Residual Module}

While the Attention Module optimizes consensus within a layer, deep architectures suffer from information loss and degradation~\cite{li2025rethinking,xie-etal-2025-rmoa}. 
To address this, we introduce the Inter-layer Residual Module for all layers $l \ge 2$, as depicted in Figure~\ref{fig:inter_layer}. 
This module mitigates the loss of critical context by constructing a residual pathway that aggregates the entire historical context.

\subsubsection{Historical Responses Accumulation}

For any layer $l \ge 2$, $\tilde{y}_l$ denotes the output generated by the current layer's Inter-agent Attention Module. The historical context stack $\mathbf{H}_l$ is defined as the sequence of outputs from all preceding layers combined with the current $\tilde{y}_l$:

$$ 
\mathbf{H}_l = [y_1, y_2, \dots, y_{l-1}, \tilde{y}_l] 
$$

\vspace{4pt}
This accumulation ensures that the system maintains a long-term memory, allowing the model to get the incremental improvements in the intermediate steps.

\subsubsection{Residual Synthesis and Propagation}

The core function of this module is to synthesize $\mathbf{H}_l$ into the final output for layer $l$. The \textbf{Residual Synthesis Agent} is denoted by $\mathcal{A}^{res}$. The output for layer $l$, is computed as:

$$ y_l = \mathcal{R}(x,\mathbf{H}_{l}, \mathcal{A}^{res}), \quad l \ge 2 $$

The Residual function $\mathcal{R}$ preserves salient information from early layers and highlights the incremental progress in $\tilde{y}_l$. 
The resulting $y_l$ is then passed to the next layer ($l+1$) as part of its input $u_{l+1}$, completing the iterative loop.

\subsubsection{Adaptive Early Stopping Mechanism}

While increased inference depth enhances reasoning capabilities, not all queries require the maximum depth $L$ to reach a satisfactory solution. 
Redundant iterations not only incur unnecessary token costs but may also lead to over-reasoning.
To optimize computational efficiency and reduce inference latency, we introduce an adaptive early stopping mechanism embedded within the Inter-layer Residual Module. 

In this enhanced setup, the Residual Synthesis Agent $\mathcal{A}^{res}$ assumes a dual responsibility: 
context synthesis and termination assessment. 
Upon generating the synthesized output $y_l$, the agent evaluates the convergence and completeness of the current solution based on the historical context $\mathbf{H}_l$.

Formally, the operation of the Residual Synthesis Agent is expanded to output a tuple containing both the synthesized response and a binary termination signal $s_l$:

$$ (y_l, s_l) = \mathcal{R}(x,\mathbf{H}_{l}, \mathcal{A}^{res}), \quad s_l \in \{0, 1\} $$

The termination signal $s_l$ is determined by analyzing the stability of the reasoning trajectory (i.e., whether $y_l$ offers significant information gain over $y_{l-1}$) and the logical completeness of the answer with respect to the initial query $x$. The inference flow is controlled as follows:

\begin{itemize}
    \item \textbf{Continue ($s_l = 0$):} If the agent determines that the current response is insufficient or unstable, the process proceeds to layer $l+1$, passing $y_l$ as part of the input $u_{l+1}$.
    \vspace{4pt}
    \item \textbf{Terminate ($s_l = 1$):} If the agent concludes that $y_l$ has reached a high confidence threshold or logical consensus, the iteration stops immediately. The current $y_l$ is returned as the final system output, bypassing all subsequent layers $k > l$.
\end{itemize}

This mechanism ensures that the system dynamically allocates computational resources, scaling the depth of reasoning proportional to the complexity of the user query.

%% file: sections/experiment.tex

\begin{table*}[tb] 
    \centering
    
    \begin{minipage}[t]{0.505\linewidth}
        \centering
        \resizebox{\linewidth}{!}{ 
\begin{tabular}{cccccc}
\toprule
\multicolumn{1}{c}{\multirow{2}{*}{\textbf{Model}}} & \multicolumn{2}{c}{\textbf{AlpacaEval 2.0}} & \multicolumn{3}{c}{\textbf{MT-Bench}} \\ 
\cmidrule(l{2pt}r{2pt}){2-6}
\multicolumn{1}{c}{} & \textbf{LC Win.} & \textbf{Win.} & \textbf{T.1} & \textbf{T.2} & \textbf{Avg.} \\ 
\midrule
Claude-4.5-Sonnet & \underline{73.49} & 61.74 & \underline{8.97} & \underline{8.28} & \underline{8.62} \\
GPT-4.1 & 69.83 & 57.23 & 8.90 & \underline{8.28} & 8.59 \\
Gemini-2.5-Pro & 65.74 & 83.02 & 8.74 & 7.99 & 8.36 \\
Qwen-Max & 64.68 & 77.22 & 8.95 & 8.16 & 8.56 \\
DeepSeek-V3.1 & 68.83 & \underline{84.02} & 8.96 & 8.16 & 8.56 \\ 
\midrule
MoA & 88.56 & 93.09 & 9.53 & 8.73 & 9.13 \\
RMoA & 78.20 & 78.48 & 9.00 & 8.64 & 8.82 \\
\textbf{Ours} & \textbf{91.15} & \textbf{95.87} & \textbf{9.60} & \textbf{9.04} & \textbf{9.32} \\
\bottomrule
\end{tabular}%
}
\caption{Performance comparison on AlpacaEval 2.0 and MT-Bench. The upper section lists individual models, while the lower section presents MoA-based methods. The best results among individual models are \underline{underlined}, and the best results among all methods are \textbf{bolded}. (T.1: Turn 1, T.2: Turn 2)}
\label{tab:main_results_large}
    \end{minipage}
    \hfill 
    \begin{minipage}[t]{0.48\linewidth}
        \centering
        \resizebox{\linewidth}{!}{ 
\begin{tabular}{cccccc}
\toprule
\multicolumn{1}{c}{\multirow{2}{*}{\textbf{Model}}} & \multicolumn{2}{c}{\textbf{AlpacaEval 2.0}} & \multicolumn{3}{c}{\textbf{MT-Bench}} \\ 
\cmidrule(l{2pt}r{2pt}){2-6}
\multicolumn{1}{c}{} & \textbf{LC Win.} & \textbf{Win.} & \textbf{T.1} & \textbf{T.2} & \textbf{Avg.} \\
\midrule

gpt-oss & 59.06 & \underline{71.82} & 8.66 & \underline{8.04} & 8.35 \\

Mistral-Small & \underline{65.94} & 65.37 & \underline{8.84} & 8.03 & \underline{8.43} \\

Qwen3 & 64.06 & 70.74 & 8.81 & 7.69 & 8.25 \\

gemma-3 & 50.60 & 65.59 & 8.75 & 7.75 & 8.25 \\

Llama-4-Scout & 44.41 & 46.27 & 8.65 & 7.26 & 7.96 \\ 

\midrule
MoA-Small & 75.07 & 86.64 & 8.93 & 8.25 & 8.59 \\
RMoA-Small & 75.79 & 87.33 & 9.01 & 8.22 & 8.62 \\
\textbf{Ours-Small} & \textbf{77.36} & \textbf{89.57} & \textbf{9.21} & \textbf{8.44} & \textbf{8.83} \\
\bottomrule
\end{tabular}
}
\caption{Performance comparison on AlpacaEval 2.0 and MT-Bench using Small-Scale Models. Attention-MoA-Small demonstrates that the architecture is effective even with smaller constituent agents.}
\label{tab:main_results_small}
    \end{minipage}
\end{table*}

\section{Experiments}

In this section, we conduct comprehensive experiments to evaluate the effectiveness of the Attention-MoA framework. We aim to answer the following research questions:

\begin{itemize}
    \item \textbf{RQ1:} Does Attention-MoA outperform existing state-of-the-art (SOTA) models and other MoA-based architectures across different evaluation benchmarks?
    \vspace{4pt}
    \item \textbf{RQ2:} Does the Attention-MoA framework demonstrate generalizability across different model scales, particularly when applied to smaller open-source models?
    \vspace{4pt}
    \item \textbf{RQ3:} How does the number of Collaborative Agents impact the overall performance?
    \vspace{4pt}
    \item \textbf{RQ4:} To what extent does the capability of the Aggregation Agent (Intra-layer Summary and Residual Synthesis) influence the overall effectiveness of the framework?
    \vspace{4pt}
    \item \textbf{RQ5:} How does the depth of the framework (number of layers) influence the generation quality?
    \vspace{4pt}
    \item \textbf{RQ6:} How does the Adaptive Early Stopping mechanism balance inference efficiency with performance?
\end{itemize}

\subsection{Experimental Setup}

\subsubsection{Datasets and Metrics}
To ensure a holistic evaluation, we employ three diverse benchmarks: \textbf{AlpacaEval 2.0}~\cite{dubois2024length}, \textbf{MT-Bench}~\cite{zheng2023judging}, and \textbf{FLASK}~\cite{ye2024flask}.

\begin{itemize}
    \item \textbf{AlpacaEval 2.0:} We utilize this benchmark to assess the macroscopic performance of our method across a wide range of user queries. Crucially, we focus on the Length-Controlled (LC) Win Rate. As MoA-based architectures (including MoA, RMoA, and Attention-MoA) tend to generate longer responses as depth increases, traditional metrics may exhibit a length bias. The LC Win Rate neutralizes this factor, providing a fairer comparison of response quality. 
   \vspace{4pt}
    \item \textbf{FLASK:} To quantitatively evaluate fine-grained capabilities, we select the \textit{Hard} subset of the FLASK dataset (89 instances). This benchmark assesses performance across 12 specific dimensions, covering a broad spectrum of capabilities ranging from logical reasoning and background knowledge to user alignment.
   \vspace{4pt}
    \item \textbf{MT-Bench:} This dataset is employed to evaluate the model's performance in multi-turn dialogues across eight diverse categories. It tests the ability to maintain context and follow instructions in domains ranging from creative generation to reasoning-intensive tasks.
\end{itemize}

\subsubsection{Model Configuration}
We evaluate Attention-MoA using two distinct configurations based on model size:

\begin{itemize}
  \item \textbf{Large-Scale Configuration:}
  This setup utilizes SOTA large language models as Collaborative Agents: \textit{Claude-4.5-Sonnet}~\cite{anthropic2025claude45}, \textit{Gemini-2.5-Pro}~\cite{comanici2025gemini}, \textit{GPT-4.1}~\cite{openai2025gpt41}, \textit{Qwen-Max}~\cite{qwen3max}, and \textit{DeepSeek-V3.1}~\cite{liu2024deepseek}. For the Intra-layer Summary Agent and Residual Synthesis Agent, we employ \textit{Claude-4.5-Sonnet}.
\vspace{4pt}
  \item \textbf{Small-Scale Configuration:}
  This setup tests the framework on smaller, efficient models: \textit{Mistral-Small-3.2-24B-Instruct-2506}~\cite{rastogi2025magistral}, \textit{Qwen3-32B}~\cite{yang2025qwen3}, \textit{gemma-3-12b-it}~\cite{team2025gemma}, \textit{Llama-4-Scout-17B-16E-Instruct}~\cite{meta2025llama4}, and \textit{gpt-oss-20b}~\cite{agarwal2025gpt}. Here, \textit{gpt-oss-20b} serves as the aggregation agent.
\end{itemize}

\subsubsection{Baselines}
We compare our method against two categories of baselines:

\begin{itemize}
    \item \textbf{Individual Models:} We evaluate all LLMs referred above, running independently to establish a performance baseline for isolated capabilities without collaboration.
    \vspace{4pt}
    \item \textbf{MoA-based Architectures:} To validate the specific contributions of our attention and residual mechanisms, we compare against existing collaborative frameworks:
   \vspace{4pt}
    \begin{itemize}
        \item Mixture-of-Agents (MoA): The standard layered collaboration framework that aggregates all reference agent outputs, relying on direct concatenation.
        \vspace{4pt}
        \item Residual MoA (RMoA): A state-of-the-art variant that optimizes collaboration through Greedy Diversity Embedding Selection to maximize agent heterogeneity and incorporates Residual Agents to mitigate inter-layer information loss.
    \end{itemize}
\end{itemize}

For a fair comparison, all MoA-based baselines utilize the exact same set of reference models (Collaborative Agents) and the same aggregation model as our Attention-MoA configuration in their respective size categories.

\begin{figure*}[t] 
  \centering
  \begin{subfigure}{0.48\linewidth} 
    \centering
    \includegraphics[width=\linewidth]{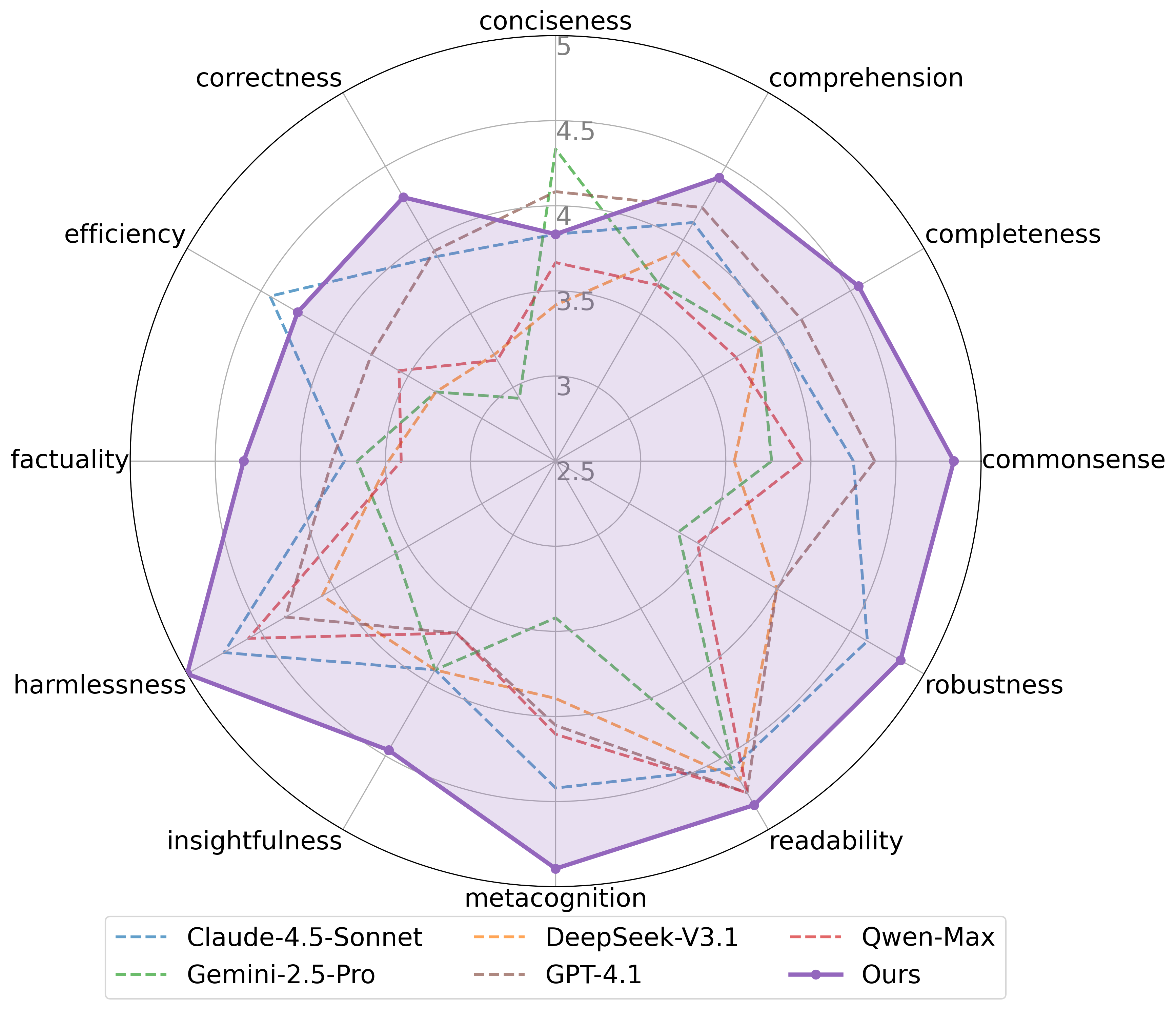}
    \caption{Comparison with Individual LLMs}
    \label{fig:flask_single}
  \end{subfigure}
  \hfill 
  \begin{subfigure}{0.48\linewidth} 
    \centering
    \includegraphics[width=\linewidth]{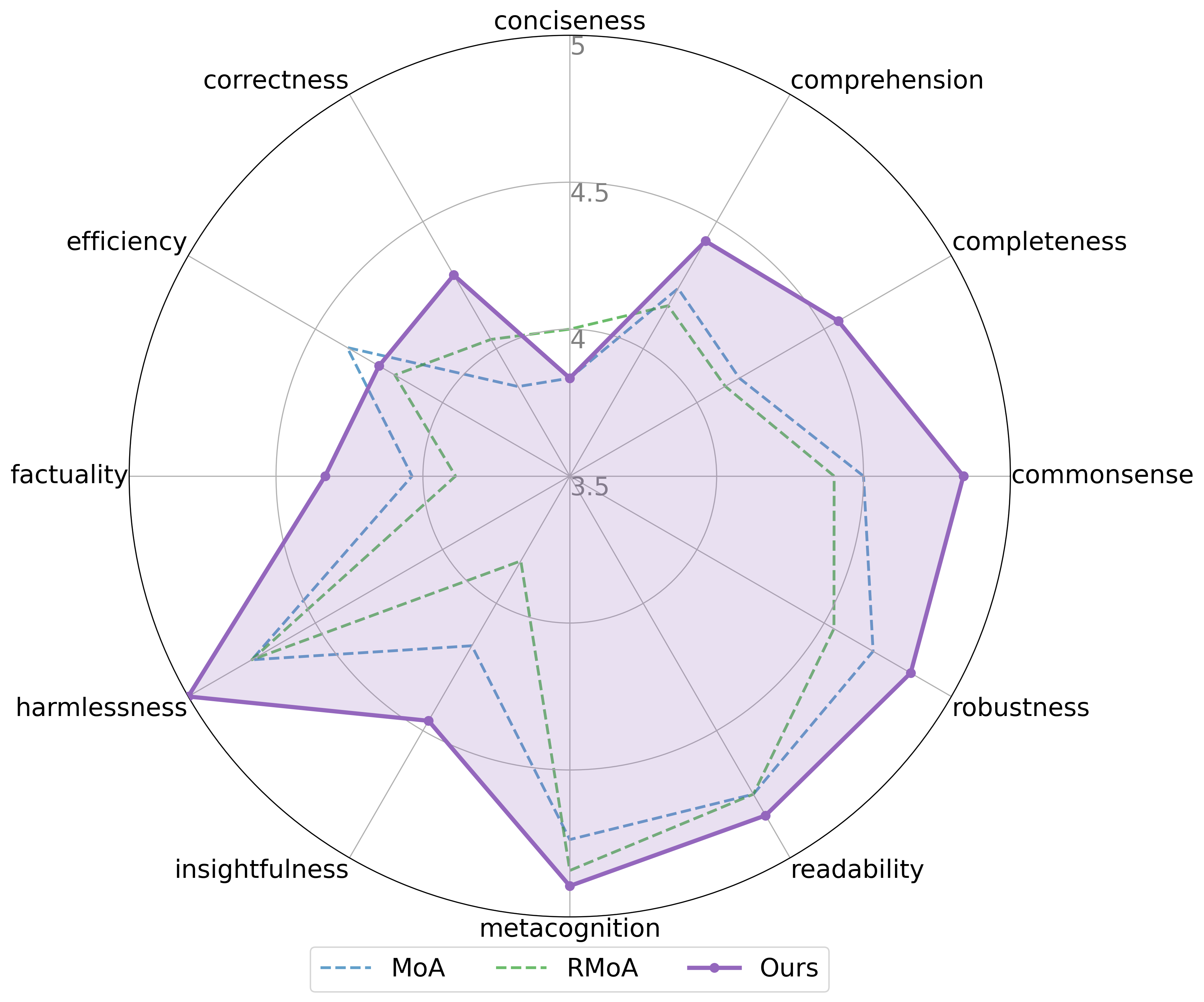}
    \vspace{-5pt}
    \caption{Comparison with MoA and RMoA Baselines}
    \label{fig:flask_moa}
  \end{subfigure}
  \caption{Fine-grained evaluation on the FLASK dataset (Hard subset). Subfigure (a) demonstrates the performance gain of Attention-MoA over individual constituent models, while (b) highlights the improvements compared to other MoA-based architectures.}
  \label{fig:flask_results}
\end{figure*}

\begin{figure*}[tb]
  \centering
  \begin{subfigure}{0.48\linewidth}
    \centering
    \includegraphics[width=\linewidth]{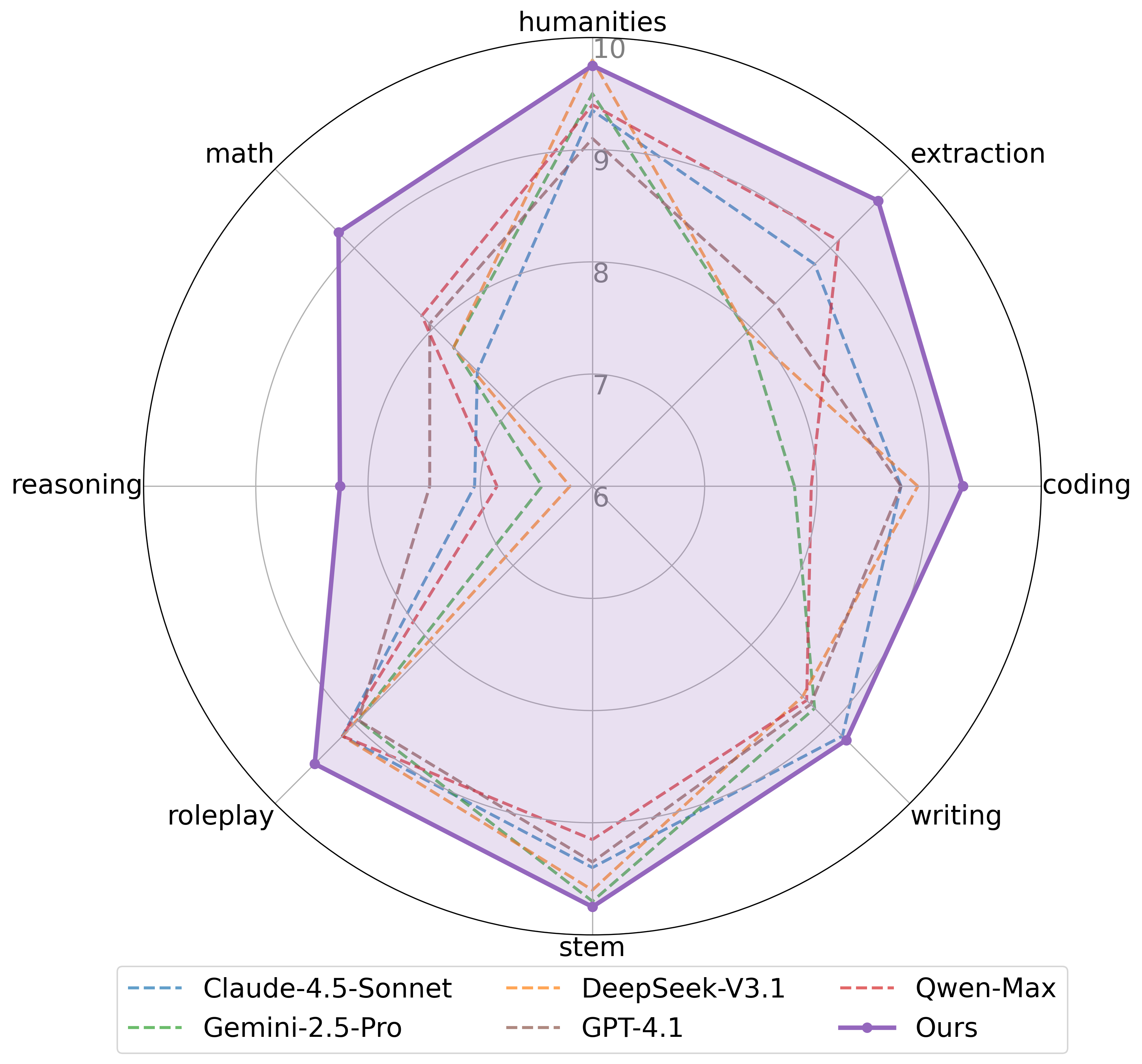}
    \caption{Comparison with Individual LLMs}
    \label{fig:mtbench_single}
  \end{subfigure}
  \hfill
  \begin{subfigure}{0.48\linewidth}
    \centering
    \includegraphics[width=\linewidth]{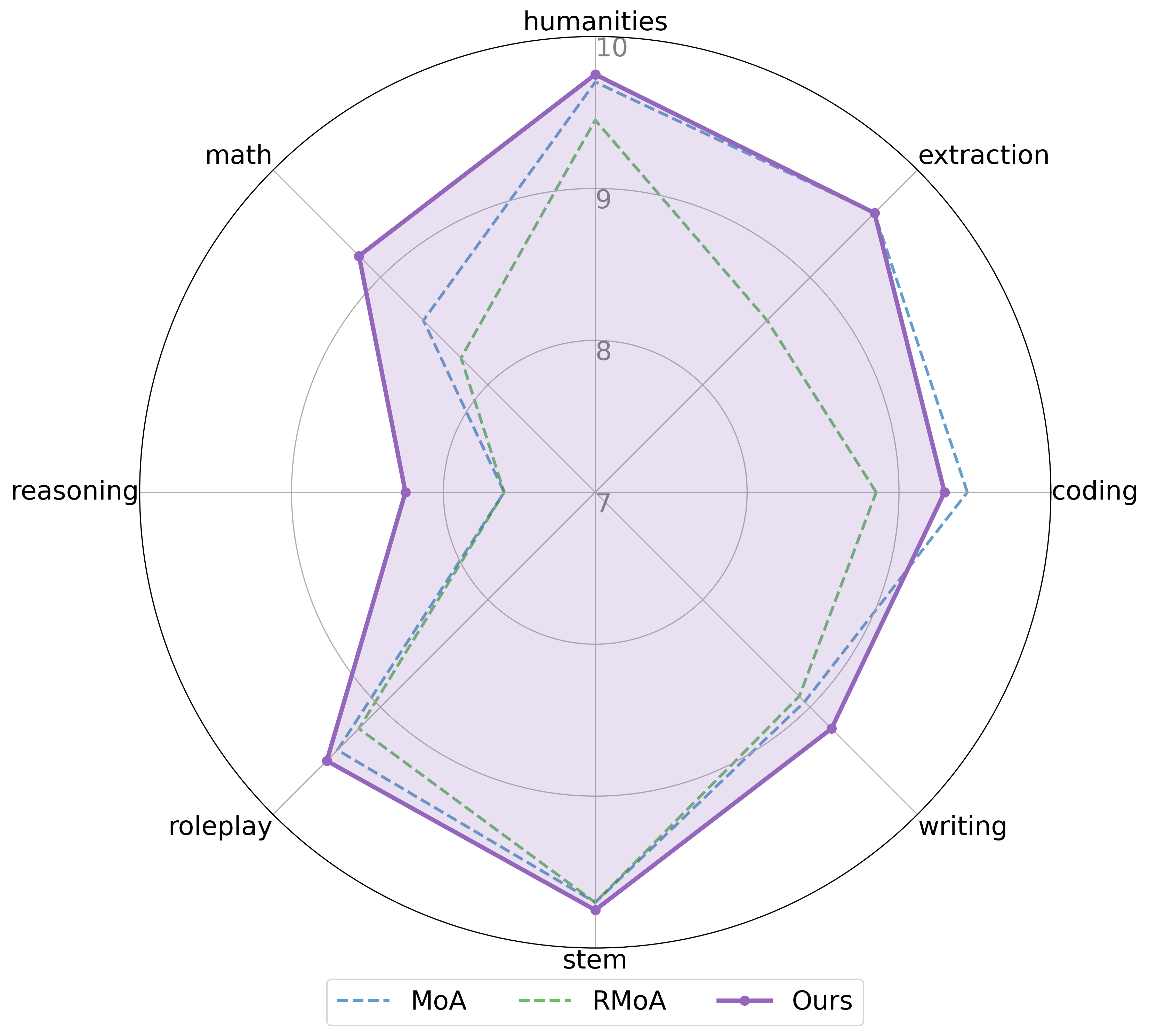}
    \vspace{-4pt}
    \caption{Comparison with MoA and RMoA Baselines}
    \label{fig:mtbench_moa}
  \end{subfigure}
\caption{Category-wise performance breakdown on MT-Bench. Subfigure (a) illustrates the comparison with individual models, while (b) shows the comparison against MoA-based baselines.}
  \label{fig:mtbench_breakdown}
\end{figure*}

\begin{figure}[tb]
  \centering
  \includegraphics[width=0.8\linewidth]{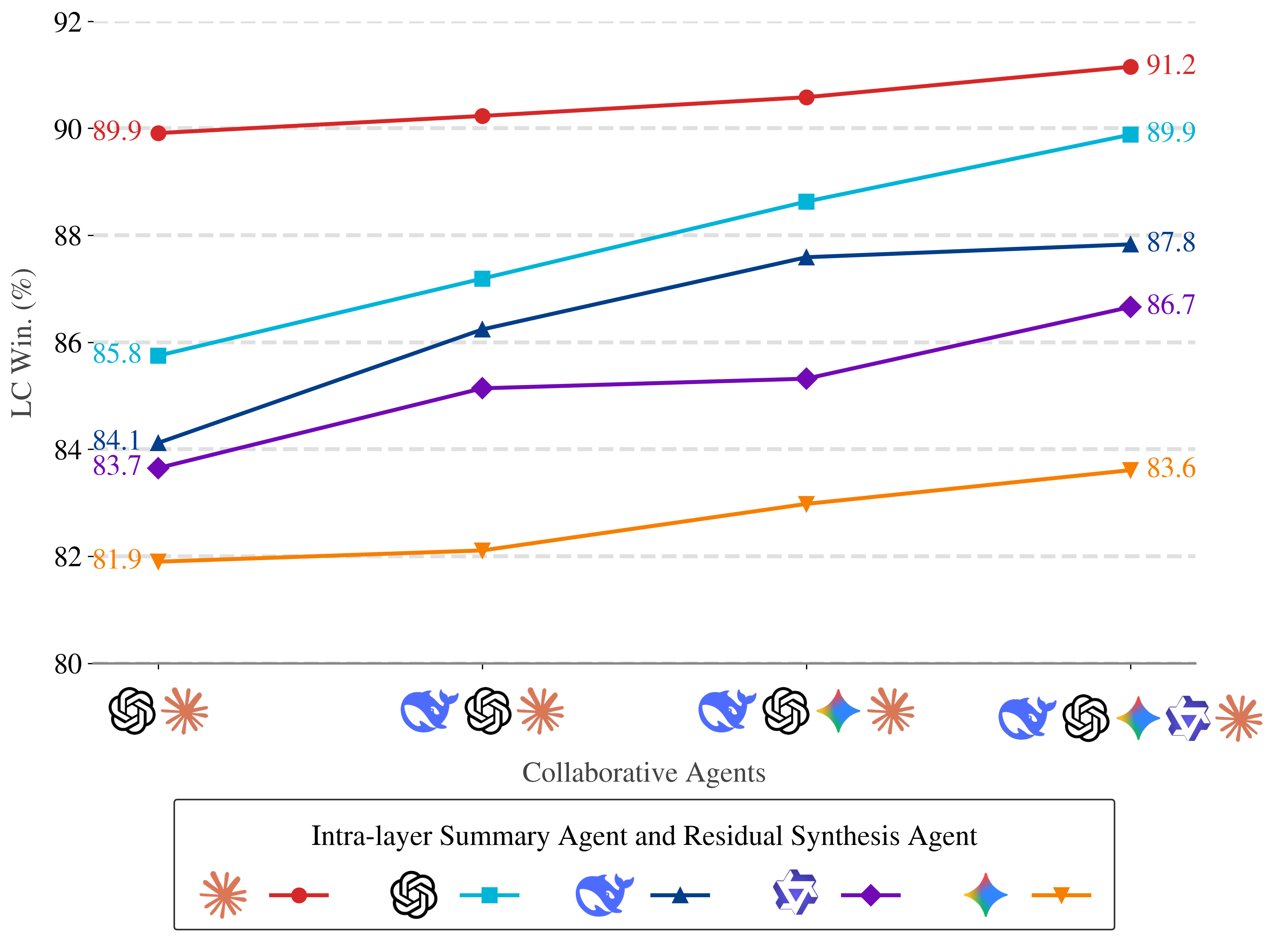}
  \caption{Impact of the number of Collaborative Agents and the capability of the Aggregation Agent on AlpacaEval 2.0 LC Win Rate. 
  The x-axis represents the set of collaborative agents, increasing in size from left to right. }
  \label{fig:ablation_study}
\end{figure}
\subsection{Main Results (RQ1)}

We first analyze the performance of Attention-MoA (Ours) using the Large-Scale configuration. 
For all MoA-based architectures (including MoA, RMoA, and Ours), we set the maximum iteration depth to $L=5$ and report the performance of the best-performing layer.
The comparative results on AlpacaEval 2.0 and MT-Bench are presented in Table \ref{tab:main_results_large}.

As shown in Table \ref{tab:main_results_large}, Attention-MoA significantly outperforms both individual SOTA models and existing MoA architectures. 
Notably, on AlpacaEval 2.0, our method achieves a Length-Controlled Win Rate of 91.15\%, surpassing the peak performance of MoA and RMoA by 2.59\% and 12.95\%, respectively. 
This indicates that the improvements are driven by higher response quality rather than mere verbosity. 
In MT-Bench, Attention-MoA demonstrates superior capability in multi-turn reasoning, achieving an average score of 9.32, with a notable improvement in the second turn (9.04) compared to the best records of MoA (8.73) and RMoA (8.64).

Figure~\ref{fig:flask_results} illustrates the fine-grained evaluation on the FLASK dataset. Attention-MoA achieves superior performance in 10 out of 12 dimensions, demonstrating a robust enhancement in overall capability. 

We observe a slight trade-off in metrics favoring brevity. Specifically, for Conciseness, our method trails behind specific single models (e.g., Gemini-2.5-Pro, GPT-4.1) and RMoA; similarly, for \textit{Efficiency}, it is surpassed by Claude-4.5-Sonnet and the standard MoA. This is an expected consequence of the Attention-MoA architecture: by synthesizing diverse perspectives to maximize information density, the model tends to generate more comprehensive and detailed responses, which naturally impacts brevity-based scores.

However, this trade-off yields substantial gains in critical cognitive and safety dimensions. As shown in the radar charts, Attention-MoA achieves a dominant lead in \textit{Harmlessness}, \textit{Factuality}, \textit{Insightfulness}, \textit{Metacognition}, \textit{Commonsense}, and \textit{Completeness}, significantly outperforming both SOTA individual models and other MoA-based architectures. This result strongly validates our hypothesis: our Inter-agent Semantic Attention Module effectively functions as a peer-review mechanism, filtering out hallucinations and safety risks while aggregating deep insights, thus prioritizing high-quality, reliable content over simple brevity.

To provide a more granular understanding of the model's capabilities, we further dissect the performance on MT-Bench across its eight specific categories: \textit{Writing}, \textit{Roleplay}, \textit{Reasoning}, \textit{Math}, \textit{Coding}, \textit{Extraction}, \textit{STEM}, and \textit{Humanities}. The category-wise breakdown is visualized in Figure \ref{fig:mtbench_breakdown}.

Figure~\ref{fig:mtbench_single} reveals how Attention-MoA successfully constructs a convex-hull of capabilities by integrating the diverse strengths of heterogeneous agents. It comprehensively outperforms individual models in most dimensions while matching the peak performance of domain-specific experts in their respective strongholds. Specifically, Attention-MoA achieves parity with Claude-4.5-Sonnet in \textit{Writing}, DeepSeek-V3.1 in \textit{Humanities}, and Gemini-2.5-Pro in \textit{STEM}. This confirms that the Intra-layer Global Summarization and Residual Synthesis mechanisms effectively capture and synthesize the best attributes of each constituent model, ensuring the final output represents the collective maximum rather than an average.

Furthermore, as illustrated in Figure~\ref{fig:mtbench_moa}, Attention-MoA maintains a robust advantage over MoA and RMoA baselines in the majority of categories. While slightly outperformed by MoA in \textit{Coding}, our method demonstrates a pronounced superiority in abstract reasoning and profound cognitive tasks, specifically \textit{Reasoning}, \textit{Math} and \textit{Writing}.
This suggests that while simple aggregation (MoA) may competitive in tasks like \textit{Coding} and \textit{Extraction} which mainly translate user intent into procedural syntax, the Inter-agent Semantic Attention Module is indispensable for complex logical deductions, effectively reducing error propagation in multi-step and profound reasoning.

\subsection{Generalizability Across Model Scales (RQ2)}

To verify the generalizability of our framework, we evaluate its performance using the Small-Scale configuration. The results are detailed in Table \ref{tab:main_results_small}.

The results in Table \ref{tab:main_results_small} confirm that Attention-MoA generalizes well to smaller models. Attention-MoA-Small (Ours-Small) achieves an MT-Bench average of 8.83 and an AlpacaEval LC Win Rate of 77.36\%. 

A compelling finding is that the small configuration of Attention-MoA rivals or even surpasses the performance of individual large commercial LLMs. For instance, Attention-MoA-Small (MT-Bench 8.83) outperforms Claude-4.5-Sonnet (8.62) and GPT-4.1 (8.59). This highlights the efficiency of our framework: 
by orchestrating a group of smaller, cost-effective models via attention and residual mechanisms, we can achieve performance levels typically reserved for significantly larger, closed-source models.

\subsection{Impact of Collaborative Agent (RQ3)}

To investigate how the scale of collaboration affects performance, we conducted an ablation by varying the number of Collaborative Agents from 2 to 5, as visualized in Figure \ref{fig:ablation_study}.

A consistent positive correlation is observed between the number of Collaborative Agents and the overall system performance (LC Win Rate).
This trend holds true regardless of which model is used as the aggregation agent. 

This finding answers RQ3, demonstrating that Attention-MoA effectively scales with the number of agents. 
The framework successfully synthesizes the diverse insights provided by additional agents, converting the increased breadth of information into higher-quality responses. 
It suggests that the Inter-agent Semantic Attention mechanism is robust enough to handle the increased complexity and potential noise introduced by more agents.

\subsection{Impact of Aggregation Agent Capability (RQ4)}

To address RQ4, we analyze the impact of the Aggregation Agent. 
We treat the Intra-layer Summary Agent and the Residual Synthesis Agent as a unified entity in this ablation study, using the same model for both. 
This choice reflects their identical functional role: synthesizing multiple input streams into a single, high-quality summary.

Figure \ref{fig:ablation_study} provides critical insights into the role of the Aggregation Agent. The results reveal a significant performance stratification based on the aggregator's choice, yet this does not strictly align with individual generation capabilities.
The configuration utilizing Claude-4.5-Sonnet consistently outperforms others by a wide margin, demonstrating it as the most effective aggregator.
Even with the maximum number of collaborative agents, the gap between the best  (91.15\%) and the weakest one (78.33\%) remains 12.82\%.
While Gemini-2.5-Pro exhibits strong individual performance on AlpacaEval 2.0, it surprisingly yields the poorest performance when serving as an Aggregation Agent.

This suggests that aggregation requires a distinct set of competencies beyond standard text generation, such as, \textit{Long-Context Reasoning} and \textit{Conflict Resolution ability}.

\begin{figure}[tb]
\centering
\includegraphics[width=0.8\linewidth]{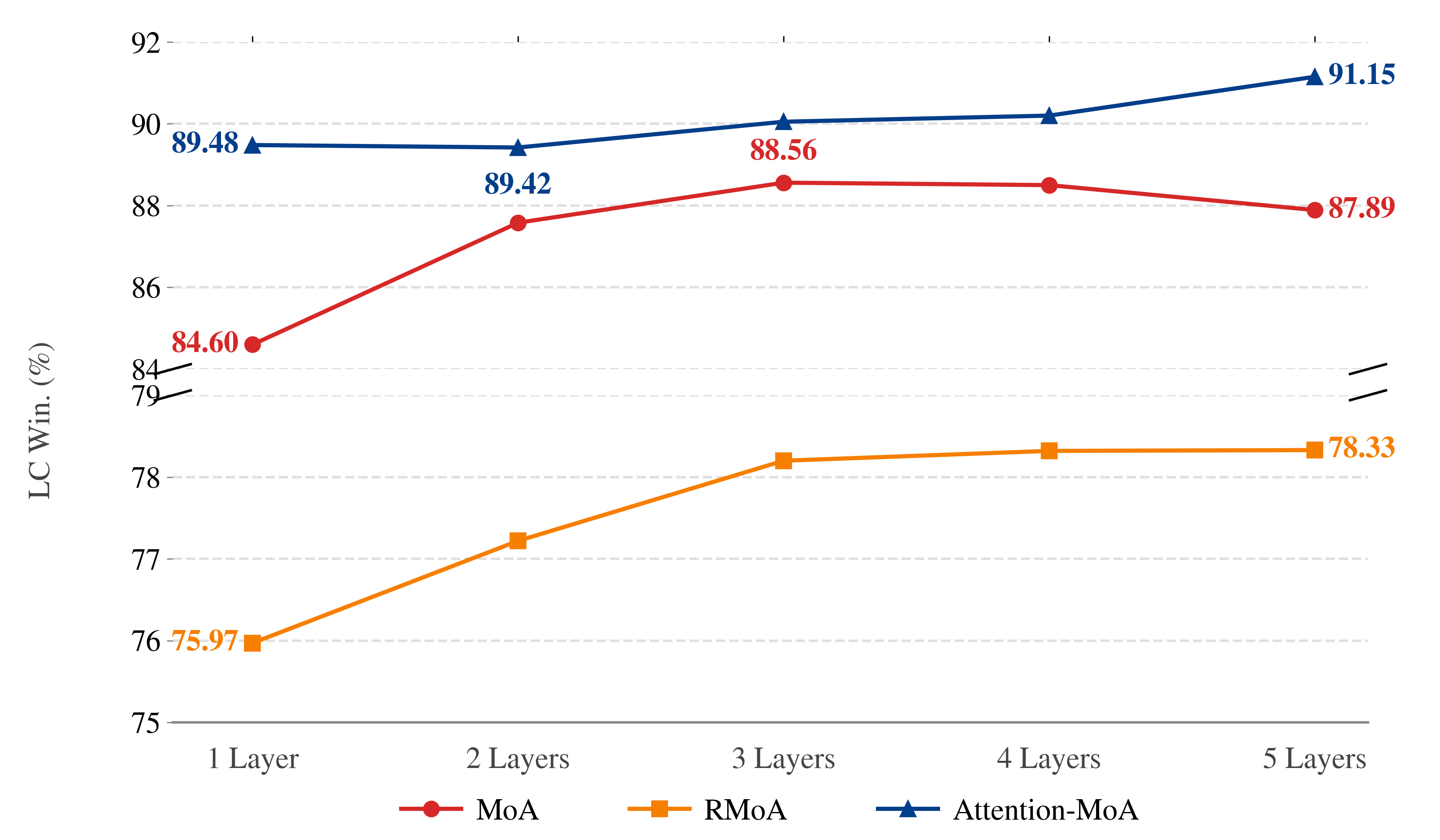} 
\caption{Performance comparison with other MoA-based baselines on AlpacaEval 2.0 across different layer depths.}

\label{fig:layer_impact}
\end{figure}

\begin{figure}[tb]
\centering
\includegraphics[width=0.8\linewidth]{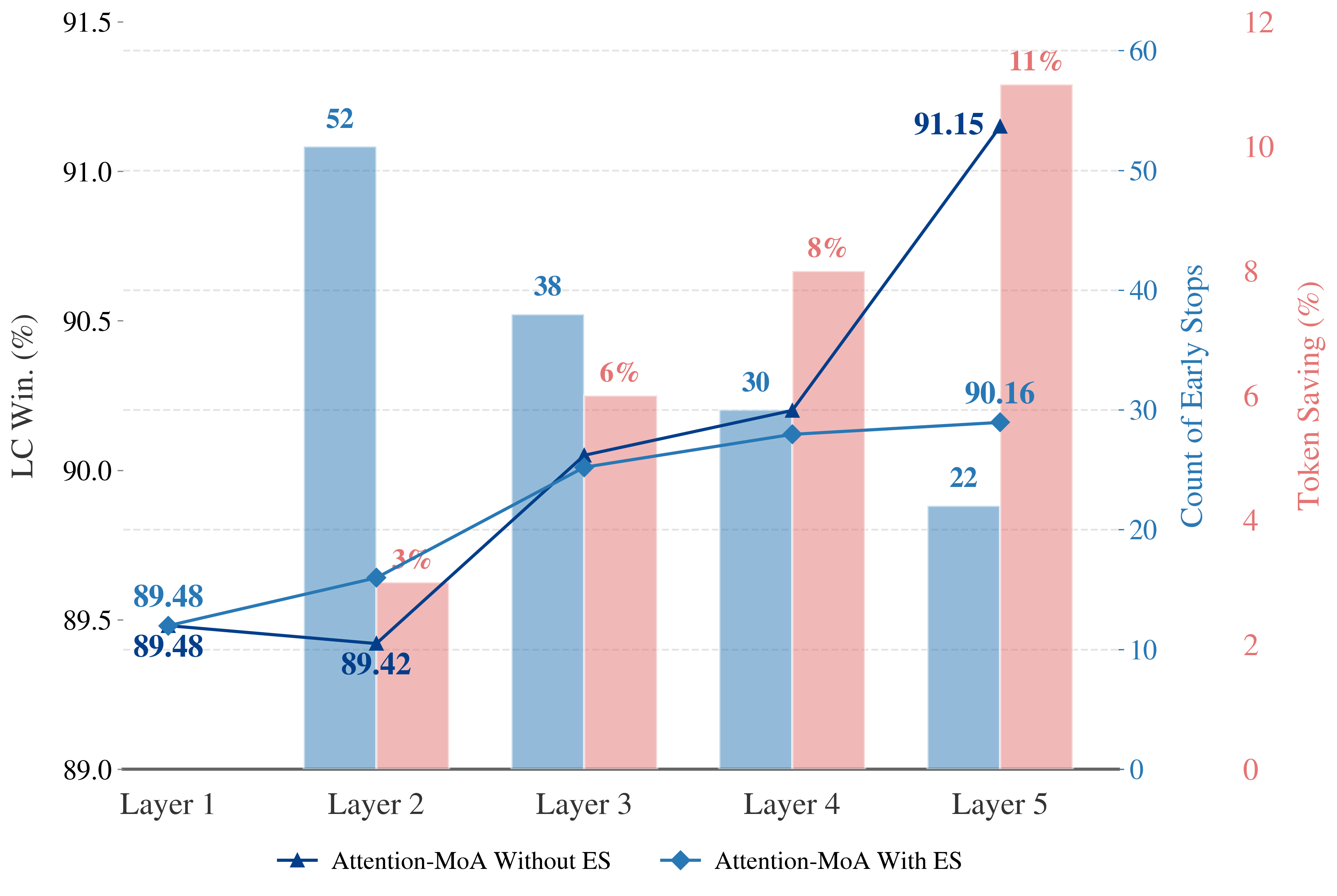}
\caption{Layer-wise analysis of Adaptive Early Stopping. 
Left y-axis: Comparison of LC Win. with and without ES. 
Right y-axes: The blue bars indicate the count of early stops, and the red bars highlights the corresponding Token Saving at different depth.}
\label{fig:early_stopping_dist}
\end{figure}
\subsection{Influence of Layer Depth (RQ5)}
To answer RQ5, we investigate how the depth of Attention-MoA influences response quality. 
We varied the number of layers from 1 to 5 and recorded the AlpacaEval LC Win Rate. 

As shown in Figure \ref{fig:layer_impact}, Attention-MoA exhibits a monotonically increasing trend in performance as the number of layers grows. 
Starting from a strong baseline of 89.48\% at Layer 1, the performance steadily climbs to 91.15\% at Layer 5.

Conversely, the baselines exhibit distinct limitations regarding depth scaling. 
MoA displays a peaking phenomenon; while it gains performance up to layer 3 (88.56\%), it suffers from degradation in subsequent layers, dropping to 87.89\% due to error accumulation. 
Meanwhile, although RMoA avoids this degradation via residual connections, it fails to achieve further growth, plateauing at 78.33\% after layer 3. 

This result shows the sustainable scalability of Attention-MoA. 
Distinct from existing methods that suffer from performance saturation or information degradation, our framework maintains a positive correlation between depth and quality. 
This indicates that Attention-MoA effectively convert increased computational resources into continuous, tangible performance gains.

\subsection{Efficiency vs. Quality Analysis (RQ6)}

To address RQ6, we evaluate the impact of the Adaptive Early Stopping (ES) mechanism, as shown in Figure \ref{fig:early_stopping_dist}.

The ES mechanism significantly optimizes resource utilization. Quantitatively, it reduces the average total tokens per query from 118,9k to 106,1k, achieving about 11\% reduction in inference cost when using prefix-caching and deepest layer is set as 5.
Interestingly, while ES leads to a plateau at deeper layers, it enhances performance at shallower depths (Layers 2-3) by terminating responses that cause quality degradation.
Prefix-Caching is applied during the Inter-Agent Semantic Attention Calculation and Residual Synthesis and Propagation  to minimize redundant computations for shared contexts.
More ablation and comparison of efficiency and cost details are provided in the \textit{Supplementary Materials}.

A noteworthy observation is that at shallower depths (Layer 2), the absence of the ES mechanism leads to unstable performance fluctuations, when computational resources are constrained. 
However, when sufficient resources allow for deeper optimization (exceeding Layer 4), the residual mechanism significantly improve the performance, converting increasing depth into performance scalability.

%% file: sections/conclusion.tex
\section{Conclusion}

In this work, we introduced Attention-MoA, a framework that enhances multi-agent collaboration through Inter-agent Semantic Attention and Inter-layer Residual Connections. 
By effectively mitigate hallucination and information degradation, our method outperforms SOTA models (e.g., Claud-4.5-sonnet) and existing MoA architectures across AlpacaEval 2.0, MT-Bench, and FLASK. 
Furthermore, we demonstrate that ensembles of small open-source models can rival massive ones, offering a scalable path for high-performance LLMs. 
While the Adaptive Early Stopping mechanism currently improves efficiency, future work will focus on further optimizing computational costs while maintaining scalability.

%% file: sections/appendix.tex
\setcounter{figure}{0}
\renewcommand{\thefigure}{A\arabic{figure}}
\setcounter{table}{0}
\renewcommand{\thetable}{A\arabic{table}}
\setcounter{equation}{0}
\renewcommand{\theequation}{A\arabic{equation}}

\section*{Appendix}
In the appendix, we provide comprehensive implementation details and extended experimental analyses to further substantiate the effectiveness of the proposed \textbf{Attention-MoA}. The content is organized as follows:

\begin{itemize}
    \item \textbf{Section~\ref{sec:prompt_templates}} presents the detailed prompt templates used across the framework, including instructions for heterogeneous agent sampling, inter-agent semantic attention (self and cross-attention), and residual synthesis.
    \item \textbf{Section~\ref{sec:appendix_efficiency}} offers a granular analysis of computational efficiency. We evaluate the trade-offs between token consumption and model performance, highlighting the specific contributions of \textit{Prefix-Caching} and \textit{Adaptive Early Stopping}. \textbf{Crucially, we demonstrate that Attention-MoA consistently outperforms other MoA-based baselines when restricted to equivalent computational budgets (token consumption).}
    \item \textbf{Section~\ref{sec:quantitative_correction}} provides a quantitative assessment of the Inter-agent Semantic Attention Module, specifically measuring its success rate in correcting hallucinations and enhancing response quality in the initial layers.
    \item \textbf{Section~\ref{sec:qualitative_analysis}} details case studies from the FLASK benchmark. We compare our method with strong baselines (MoA and RMoA) in domains requiring strict logical reasoning (Chess) and complex information synthesis (Humanities).
    \item \textbf{Section~\ref{sec:ai_usage}} provides a declaration of AI tool usage, detailing their role in manuscript refinement and the verification of complex case studies in compliance with conference guidelines.
    \item \textbf{Section~\ref{sec:conclusion}} concludes the material by summarizing the findings regarding the framework's implementation, efficiency gains, and superior reasoning capabilities.
\end{itemize}

\section{Prompt Templates}
\label{sec:prompt_templates}

We organize the prompt templates according to the two primary modules of the Attention-MoA framework: the \textbf{Inter-agent Semantic Attention Module} and the \textbf{Inter-layer Residual Module}.
\subsection{Inter-agent Semantic Attention Module}
This module orchestrates the intra-layer collaboration and refinement process. It utilizes the following templates:
\begin{itemize}
    \item \textbf{Heterogeneous Agent Response Sampling Strategy (Figure \ref{fig:prompt_heterogeneous}):} This template is employed in the initial phase of each layer to guide diverse agents to generate independent responses based on the user query and conversation history.
    \item \textbf{Inter-Agent Semantic Attention Calculation (Figure \ref{fig:prompt_attention}):} This set of templates facilitates the critique-and-refine process. It includes \textit{Self-Attention}, \textit{Cross-Attention} and \textit{Semantic Attention Aggregation} for agents interaction. 
    Notably, for the \textit{Cross-Attention} component, we provide two variants to balance precision and efficiency: a pairwise version ($\mathcal{O}(N^2)$) and a single-pass version ($\mathcal{O}(N)$).
    \item \textbf{Intra-layer Summarization (Figure \ref{fig:prompt_summarization}):} This template instructs the Summary Agent to take the suggestions from all agents within a layer and refine its initial response.
\end{itemize}

\subsection{Inter-layer Residual Module}
The \textbf{Residual Synthesis and Propagation (Figure \ref{fig:prompt_res})} manages the information flow and termination logic across layers. 
We provide two versions: one incorporating the Adaptive Early Stopping mechanism (which includes specific instructions to output a signal upon convergence) and a standard version without Early Stopping for fixed-depth inference and scalable performance.

\begin{table*}[tb]
\centering
\begin{tabular}{cccccc}
\toprule
Methods        & Prefix-Caching & ES & Layer & Tokens & LC Win. \\
\midrule
RMoA           & -              & -  & 1     & 32k    & 75.97   \\
RMoA           & -              & -  & 2     & 105k   & 77.22   \\
RMoA           & -              & -  & 3     & 164k   & 78.20   \\
MoA            & -              & -  & 1     & 44k    & 84.60   \\
MoA            & -              & -  & 2     & 161k   & 87.58   \\
\midrule
RMoA           & -              & -  & 4     & 220k   & 78.32   \\
RMoA           & -              & -  & 5     & 287k   & 78.33   \\
MoA            & -              & -  & 3     & 287k   & 88.56   \\
\rowcolor{gray!30} Attention-MoA &\ding{51}             &\ding{51} & 1     & 204k   & 89.48   \\
\rowcolor{gray!30} Attention-MoA &\ding{51}             &\ding{55}  & 1     & 204k   & 89.48   \\
\rowcolor{gray!30} Attention-MoA &\ding{55}              &\ding{55}  & 1     & 285k   & 89.48   \\
\midrule
MoA            & -              & -  & 4     & 404k   & 88.50   \\
MoA            & -              & -  & 5     & 523k   & 87.89   \\
\rowcolor{gray!30} Attention-MoA &\ding{51}             &\ding{51} & 2     & 425k   & 89.64   \\
\rowcolor{gray!30} Attention-MoA &\ding{51}             &\ding{55}  & 2     & 442k   & 89.42   \\
\rowcolor{gray!30} Attention-MoA &\ding{51}             &\ding{51} & 3     & 600k   & 90.01   \\
\rowcolor{gray!30} Attention-MoA &\ding{55}              &\ding{55}  & 2     & 600k   & 89.42   \\

\midrule

\rowcolor{gray!30} Attention-MoA &\ding{51}             &\ding{51} & 4     & 844k   & 90.12   \\
\rowcolor{gray!30} Attention-MoA &\ding{51}             &\ding{51} & 5     & 106,1k & 90.16   \\

\rowcolor{gray!30} Attention-MoA &\ding{51}              &\ding{55}  & 3     & 641k   & 90.05   \\
\rowcolor{gray!30} Attention-MoA &\ding{51}              &\ding{55}  & 4     & 926k & 90.20   \\
\rowcolor{gray!30} Attention-MoA &\ding{51}              &\ding{55}  & 5     & 118,9k & 91.15 
\\
\bottomrule
\end{tabular}

\caption{Comparison of computational cost (Tokens) and performance (AlpacaEval 2.0 LC Win.) between Attention-MoA and other MoA-based methods. The table is divided into four sections by solid lines, corresponding to token consumption ranges of $<$200k, 200k-400k, 400k-600k, and $>$600k. The table highlights the efficiency gains from Prefix-Caching and Early Stopping (ES).}
\label{tab:cost_performance}
\end{table*}

\section{Details of Computational Efficiency and Performance}
\label{sec:appendix_efficiency}

In this section, we provide a comprehensive evaluation of the computational overhead associated with the Attention-MoA framework compared to existing multi-agent baselines, specifically MoA and RMoA. Table~\ref{tab:cost_performance} details the relationship between the number of layers, total token consumption, and the performance on AlpacaEval 2.0 (LC Win Rate).

\subsection{Comparison with Baselines}

Attention-MoA demonstrates superior scalability and resource utilization compared to RMoA and MoA:

\begin{itemize}
    \item \textbf{Superior Performance at Equivalent Computational Cost:} 
    Attention-MoA significantly outperforms baselines when restricted to similar token consumption. As shown in Table~\ref{tab:cost_performance}, with a budget of approximately 285k tokens, Attention-MoA (Layer 1) achieves an LC Win Rate of 89.48\%. In comparison, MoA requires 3 layers to consume a similar amount of resources (287k tokens) but only reaches a score of 88.56\%, while RMoA lags further behind at 78.33\% (Layer 5, 287k tokens).
    \item \textbf{Continuous Performance Gains at Depth:} 
    Unlike MoA, which suffers from performance degradation at deeper layers (dropping from 88.56\% at Layer 3 to 87.89\% at Layer 5), Attention-MoA effectively utilizes additional computational resources to drive further improvements. Even in the most computationally intensive setting (without Early Stopping), extending the model from Layer 4 to Layer 5 incurs an additional cost of about 263k tokens but yields a substantial performance boost of 0.95\%, elevating the score from 90.20\% to a peak of 91.15\%. This demonstrates that our architecture progressively optimizes performance with increasing depth, continuing to refine reasoning capabilities where other methods saturate.

\end{itemize}

In summary, the combination of the Inter-agent Semantic Attention Module with efficient memory management and adaptive depth control allows Attention-MoA to outperform existing methods significantly, delivering higher quality responses with optimized computational resources.

\subsection{Impact of Optimization Mechanisms}

Our framework incorporates two critical strategies to optimize inference efficiency: Prefix-Caching and Adaptive Early Stopping (ES).

\paragraph{Prefix-Caching Efficiency.}
As elaborated in the methodology and visualized in Figures~\ref{fig:prompt_attention} and~\ref{fig:prompt_res}, computational redundancy primarily stems from the \textbf{Inter-Agent Semantic Attention Calculation} and \textbf{Residual Synthesis and Propagation} phases. 
The results in Table~\ref{tab:cost_performance} demonstrate that Prefix-Caching significantly mitigates this computational overhead. Specifically, at Layer 1, this technique reduces token consumption by approximately 28\% (from 285k down to 204k) while maintaining identical performance. 

\paragraph{Effectiveness of Adaptive Early Stopping.}
The Adaptive Early Stopping mechanism allows the model to terminate inference when a consensus is reached, preventing unnecessary iterations.
Comparing the results for Attention-MoA with and without ES, we observe that ES consistently reduces token usage while maintaining competitive performance.
Notably, at Layer 2, enabling ES reduces tokens from 442k to 425k while achieving a slightly higher win rate (89.64\% vs. 89.42\%). This suggests that ES not only saves costs but may also prevent over-reasoning or information degradation in some cases, where residual module at shallow layers lack sufficient information, leading to performance fluctuations.

Furthermore, the experiments demonstrate that our residual mechanism effectively rectifies information degradation when layer become deeper, ensuring that performance continues to improve with depth and validate the scalability of our architecture.

\section{Effectiveness Analysis of the Inter-agent Semantic Attention Module}
\label{sec:quantitative_correction}

To verify whether the \textbf{Inter-agent Semantic Attention Module} genuinely identifies and corrects hallucinations or logical errors, we further conduct an analysis on Inter-agent Semantic Attention Module.

\subsection{Experimental Setup}
We specifically focus our analysis on the first layer of Attention-MoA, as the majority of hallucinations and logical errors originate during the initial generation phase by heterogeneous agents ($r_{i,1}$). 
Since these agents generate responses independently, their answers are most susceptible to inaccuracies compared to refined deeper layers. 
Therefore, Layer 1 make itself the ideal stage to evaluate the module's capability in identifying and rectifying these early-stage errors before they propagate.

To evaluate this, we employed \textit{Claude-4.5-Sonnet} as a judge to compare the set of refined response 
$r'_{i,1}$
against the 
initial responses 
$r_{i,1}$
for all 805 queries generated by \textit{Qwen-Max} in the AlpacaEval 2.0. 
The evaluation criteria were categorized into two primary aspects:
  
\begin{itemize}
\item \textbf{Hallucination/Error Correction:} Does 
$r'_{j,1}$
  successfully correct a factual error, logical fallacy, or hallucination that was present in the majority or a significant portion of the initial responses?
\item \textbf{Quality Enhancement:} Does 
$r'_{j,1}$
  provide a distinct quality improvement in comprehensiveness, structure, or clarity compared to the initial response, even if no obvious factual errors were present?
\end{itemize}

\subsection{Results and Analysis}

Table~\ref{tab:correction_stats} summarizes the results of this evaluation. Out of 805 queries, the attention mechanism demonstrates a significant positive improvement in approximately 19\% of cases.

Most notably, we identified 46 instances (5.7\%) where the model successfully corrected explicit errors or hallucinations. 
This indicates that the Inter-agent Semantic Attention Module effectively  leverages cross-agent critique to identify inconsistencies that the initial responses overlook.

Furthermore, in 107 instances (13.3\%), the module synthesized a response of higher quality than the initial one. This validates the effectiveness of our critique-and-refine paradigm in both error mitigation and quality elevation.

\begin{table}[tb]
\centering
\resizebox{0.8\linewidth}{!}{
\begin{tabular}{lccc}
\toprule
\textbf{Metric} & \textbf{Count} & \textbf{Rate (\%)} & \textbf{Description} \\
\midrule
Total Queries & 805 & 100 & Full evaluation set \\
\midrule
\rowcolor{green!10} \textbf{Error Correction} & \textbf{46} & \textbf{5.7} & \textit{Corrected hallucinations or logical errors} \\
Quality Enhancement & 107 & 13.3 & \textit{Improved quality} \\
\midrule
Total Improved & 153 & 19.0 & Error correction + quality enhancement \\
\bottomrule
\end{tabular}
}
\vspace{5pt}
\caption{
Statistics of improvements observed in Layer~1 of Attention-MoA Inter-agent Semantic Attention Module.
}
\label{tab:correction_stats}
\end{table}

\section{Qualitative Analysis with Case Study}
\label{sec:qualitative_analysis}

To further understand the performance improvements demonstrated by Attention-MoA, we conduct a qualitative analysis using case studies from the FLASK benchmark. We selected two distinct domains—Logical Reasoning (Chess) and Humanities (Philosophy)—to illustrate how our model handles hallucination mitigation and information synthesis.

\subsection{Logical Reasoning and Error Correction}
Figure~\ref{fig:case_study_flask} presents a complex chess problem requiring precise spatial reasoning and rule adherence to find a \textit{checkmate-in-one} move. 
\begin{itemize}
    \item \textbf{Baseline -- MoA:} The standard MoA method (Blue box) hallucinates an illegal move (\textit{Qg1+}). While the reasoning sounds plausible in natural language, it fails to account for the board state constraint (the Bishop on f1 blocks the path), leading to low evaluation scores (\textit{Logical correctness}: 1, \textit{Commonsense Understanding}: 2, \textit{Completeness}: 3).
    \item \textbf{Baseline -- RMoA:}  RMoA (Red box) successfully identifies the correct move (\textit{Qxf1\#}) through a rigorous chain-of-thought. However, its reasoning process is strictly constructive rather than discriminative. It builds a valid solution path but does not demonstrate the capability to evaluate and reject competing, plausible-sounding errors (such as the illegal \textit{Qg1+}). This suggests that while RMoA is a strong solitary solver, it lacks the comparative reasoning mechanism required to handle the conflicting information streams often present in complex multi-agent environments.
    \item \textbf{Attention-MoA:} Our method (Green box) not only identifies the correct move (\textit{Qxf1\#}) -- achieving perfect evaluation scores (\textit{Harmlessness}: 5, \textit{Completeness}: 5, \textit{Factuality}: 5) -- but also demonstrates a superior self-correction capability. Notably, the model explicitly considers the illegal move suggested by other agents (\textit{Option 2: Qg1+}), evaluates its validity ("the Bishop on f1 blocks..."), and rejects it. This confirms that Attention-MoA goes beyond simple aggregation, actively leveraging semantic discrepancies to distinguish between plausible-sounding hallucinations and grounded truths.

\end{itemize}

\subsection{Comprehensive Synthesis in Humanities}
Figure~\ref{fig:case_study_human} illustrates a query from the Humanities domain regarding Thaddeus Metz's philosophical critique of consequentialism in the context of African values. This case highlights our method's ability to integrate fragmented insights.

\begin{itemize}
    \item \textbf{Baseline -- MoA \& RMoA:} The baselines exhibit complementary deficiencies, resulting in identical sub-optimal scores (\textit{Harmlessness}: 4, \textit{Completeness}: 4, \textit{Factuality}: 4). 
    RMoA (Red box) provides a theoretical deep-dive into \textit{Relational Ethics} and \textit{Intrinsic Wrongness} but overlooks the critical \textit{Empirical Contingency} argument. 
    Conversely, MoA (Blue box) correctly identifies the \textit{Empirical Contingency} flaw (the instability of consequentialist logic) but offers a superficial treatment of the underlying African philosophical concepts. 
    Both models fail to present the full scope of Metz's critique.
    \item \textbf{Attention-MoA Synthesis (Structural Integration):} Attention-MoA (Green box) achieves perfect evaluation results (\textit{Harmlessness}: 5, \textit{Completeness}: 5, \textit{Factuality}: 5). It delivers a significantly more comprehensive and in-depth analysis compared to the baselines, addressing two critical dimensions:
    \begin{itemize}
        \item \textbf{Empirical Contingency:} It captures the logical objection regarding the fragility of arguments based on changeable facts (effectively covering the aspect missed by RMoA).
        \item \textbf{Philosophical Misalignment:} It simultaneously provides the theoretical depth regarding \textit{maximizing outcomes} vs. \textit{honoring relationships} (effectively covering the aspect missed by MoA).
    \end{itemize}
\end{itemize}

This confirms that Attention-MoA transcends the fragmented coverage of the baselines, successfully unifying logical rigor with theoretical depth to construct a response of superior structural integrity and comprehensiveness.

\section{Usage of AI Tools}
\label{sec:ai_usage}

In compliance with conference guidelines, we acknowledge the use of Google's Gemini to refine the manuscript's clarity and grammatical precision. Furthermore, AI tools were employed to assist in the preliminary analysis of complex case studies, specifically for verifying chess logic and synthesizing Thaddeus Metz's philosophical arguments. The authors have meticulously reviewed all AI-assisted content and retain full responsibility for the accuracy  of the final publication.

\section{Conclusion}
\label{sec:conclusion}

This supplementary material details the implementation of \textbf{Attention-MoA} and validates its effectiveness. Our analysis demonstrates that the framework significantly outperforms baselines under equivalent computational budgets, leveraging \textit{Prefix-Caching} and \textit{Adaptive Early Stopping} for high efficiency. Furthermore, both quantitative metrics and qualitative case studies confirm the model's superior capability in correcting hallucinations and synthesizing complex information through its semantic attention mechanism.


\begin{figure}[tb]
    \centering
    
    \begin{tcolorbox}[colback=gray!5!white, colframe=gray!75!black, title=\textbf{Prompt Template for Heterogeneous Agent Response Sampling}]
    \vspace{-5pt}
        \small 
        \textbf{System Prompt:} \\
You have been provided with a set of responses from various large language models to the latest user query. 
Your task is to synthesize these responses into a single, high-quality response. It is crucial to critically evaluate the information provided in these responses, recognizing that some of it may be biased or incorrect. Your response should not simply replicate the given answers but should offer a refined, accurate, and comprehensive reply to the instruction. Ensure your response is well-structured, coherent, and adheres to the highest standards of accuracy and reliability.\\
Responses from models:\\
Response from model 1:\\
\texttt{\{response\_of\_model\_1\}}\\
Response from ...

        \textbf{User Prompt:} \\
        \texttt{\{conv\_history\}}\\
        \texttt{\{user\_query\}}  
        \vspace{-5pt}
    \end{tcolorbox}

    \caption{The prompt template used for Heterogeneous Agent Response Sampling Strategy.}
    \label{fig:prompt_heterogeneous}
\end{figure}


\begin{figure*}[tb]
    \centering
\vspace{-10pt}
    \begin{tcolorbox}[colback=gray!5!white, colframe=gray!75!black, title=\textbf{Prompt Template for Cross-Attention ($\mathcal{O}(N^2)$)}]
    \vspace{-5pt}
        \small 
        \textbf{User Prompt:} \\
You have been provided with the answer of yours and another large language model to the latest user query.
You should compare your answer with the answer of another large language model, and offer some suggestions with reasons to the answer of another large language model, if you think there are some parts of your answer that can help the another large language model to improve its answer's quality.

The history of conversation is:\\
\texttt{\{conv\_history\}}

The latest user query is:\\
\texttt{\{user\_query\}}

The answer of another large language model is:\\
\texttt{\{model\_answer\_other\}}

The answer of yours is:\\
\texttt{\{model\_answer\_own\}}

Please give your suggestions to the answer of another large language model.
    \vspace{-5pt}
    \end{tcolorbox}
\vspace{-10pt}
    \begin{tcolorbox}[colback=gray!5!white, colframe=gray!75!black, title=\textbf{Prompt Template for Cross-Attention ($\mathcal{O}(N)$)}]
    \vspace{-5pt}
        \small 
        \textbf{User Prompt:} \\
You have been provided with the answer of yours and other large language models to the latest user query.
You should compare your answer with the answers from other large language models, and offer some suggestions with reasons to the other answers, if you think there are some parts of your answer that can help the other large language models to improve their answers' quality.

The history of conversation is:\\
\texttt{\{conv\_history\}}

The latest user query is:\\
\texttt{\{user\_query\}}

The answer of other large language models are:\\
Answer from model 1:\\
\texttt{\{model\_answer\_1\}}\\
Answer from model ...

The answer of yours is:\\
\texttt{\{model\_answer\_own\}}

The output should be a JSON object as:\\
\{\\
\enquote{suggestion\_for\_model\_1}: \enquote{your suggestions for the answer of model 1},\\
\enquote{suggestion\_for\_model\_2}: ...\\
\}

Please give your suggestions to the answer of another large language model.
    \vspace{-5pt}
    \end{tcolorbox}
\vspace{-10pt}
    \begin{tcolorbox}[colback=gray!5!white, colframe=gray!75!black, title=\textbf{Prompt Template for Self-Attention}]
        \small 
        \vspace{-5pt}
        \textbf{User Prompt:} \\
You have been provided with your answer the latest user query.
You should reassess your earlier response to identify any potentially unreasonable or incorrect content, and provide revision suggestions to improve the rationality and completeness of your answer.

The history of conversation is:\\
\texttt{\{conv\_history\}}

The latest user query is:\\
\texttt{\{user\_query\}}

Your previous answer is:\\
\texttt{\{model\_answer\_own\}}

Please give your suggestions to your previous answer.
    \vspace{-5pt}
    \end{tcolorbox}
\vspace{-10pt}
    \begin{tcolorbox}[colback=gray!5!white, colframe=gray!75!black, title=\textbf{Prompt Template for Semantic Attention Aggregation}]
        \small 
        \vspace{-5pt}
        \textbf{System Prompt:} \\
You have been provided with your previous answer to the latest user query. You also have been provided the suggestions from other large language models to your previous answer. 
You need to determine whether those suggestions are correct and reasonable. Your task is to integrate those reasonable suggestions and refine your previous answer to the latest user query.

Your previous answer is:\\
\texttt{\{model\_query\}}

Belows are the suggestions from other large language models to your previous answer:\\
Suggestions from model 1:\\
\texttt{\{suggestions\_from\_model\_1\}}

Suggestions from ...

Please integrate those reasonable suggestions and refine your previous answer to the latest user query.

        \textbf{User Prompt:} \\
        \texttt{\{conv\_history\}}\\
        \texttt{\{user\_query\}}  
        \vspace{-5pt}
    \end{tcolorbox}
    \vspace{-12pt}
    \caption{The prompt template used for Inter-Agent Semantic Attention Calculation.}
    \label{fig:prompt_attention}
\end{figure*}

\begin{figure*}[tb]
    \centering
    
    \begin{tcolorbox}[colback=gray!5!white, colframe=gray!75!black, title=\textbf{Prompt Template for Intra-layer Summarization}]
        \small 
        \vspace{-5pt}
        \textbf{System Prompt:} \\
You have been provided with your previous answer to the latest user query. You also have been provided the suggestions from other large language models to your previous answer. 

You need to determine whether those suggestions are correct and reasonable. Your task is to integrate those reasonable suggestions and refine your previous answer to the latest user query.

Your previous answer is:\\
\texttt{\{model\_query\}}

Belows are the suggestions from other large language models to your previous answer:\\
Suggestions from model 1:\\
\texttt{\{suggestions\_from\_model\_1\}}

Suggestions from ...

Please integrate those reasonable suggestions and refine your previous answer to the latest user query.

        \textbf{User Prompt:} \\
        \texttt{\{conv\_history\}}\\
        \texttt{\{user\_query\}}  
        \vspace{-5pt}
    \end{tcolorbox}
\vspace{-12pt}
    
    \caption{The prompt template used for Intra-layer Summarization.}
    \label{fig:prompt_summarization}

\end{figure*}

\begin{figure*}[tb]
    \centering
    
    \begin{tcolorbox}[colback=gray!5!white, colframe=gray!75!black, title=\textbf{Prompt Template for Residual Synthesis and Propagation (without Early Stopping)}]
        \small 
        \vspace{-5pt}
        \textbf{System Prompt:} \\
This scenario resembles a multi-round deliberation involving multiple experts. 
You are given the responses of the historical discussion rounds to the latest user's query·
Your task is to synthesize these responses into a single, high-quality response. It is crucial to critically evaluate the information provided in these responses, recognizing that some of it may be biased or incorrect. 
Your response should not simply replicate the given answers but should offer a refined, accurate, and comprehensive reply to the instruction. Ensure your response is well-structured, coherent, and adheres to the highest standards of accuracy and reliability.

Responses of historical round:\\
Response of historical round 1:\\
\texttt{\{result\_from\_round\_1\}}\\
Response of historical ...

        \textbf{User Prompt:} \\
        \texttt{\{conv\_history\}}\\
        \texttt{\{user\_query\}}  
        \vspace{-5pt}
    \end{tcolorbox}

    \begin{tcolorbox}[colback=gray!5!white, colframe=gray!75!black, title=\textbf{Prompt Template for Residual Synthesis and Propagation (with Early Stopping)}]
        \small 
        \vspace{-5pt}
        \textbf{System Prompt:} \\
This scenario resembles a multi-round deliberation involving multiple experts. 
You are given the responses of the historical discussion rounds to the latest user's query·
Your task is to synthesize these responses into a single, high-quality response. It is crucial to critically evaluate the information provided in these responses, recognizing that some of it may be biased or incorrect. 
Your response should not simply replicate the given answers but should offer a refined, accurate, and comprehensive reply to the instruction. Ensure your response is well-structured, coherent, and adheres to the highest standards of accuracy and reliability.

Responses of historical round:\\
Response of historical round 1:\\
\texttt{\{result\_from\_round\_1\}}\\
Response of historical ...

If, in your judgment, the latest round’s results exhibit no improvement over those of the preceding round, you should output **Attention-MoA should be stopped** only.

        \textbf{User Prompt:} \\
        \texttt{\{conv\_history\}}\\
        \texttt{\{user\_query\}}  
        \vspace{-5pt}
    \end{tcolorbox}
    
\vspace{-12pt}
    
    \caption{The prompt template used for Residual Synthesis and Propagation.}
    \label{fig:prompt_res}

\end{figure*}

\begin{figure*}[t]
    \centering
    \tcbset{
        width=\linewidth,      
        boxrule=0.8pt,         
        arc=2mm,               
        left=2mm, right=2mm, top=2mm, bottom=2mm, 
        fonttitle=\bfseries\small, 
        fontupper=\tiny       
    }
\vspace{-15pt}
    \begin{tcolorbox}[colback=gray!5, colframe=black!60, title=User Query]
        \vspace{-5pt}
        In the following chess position, find a checkmate-in-one move.\\
        n1. g3 d6 2. Bg2 c5 3. d3 b6 4. Nf3 Nc6 5. O-O Bb7 6. b3 f6 7. c4 e5 8. Bb2 Nge7 9. d4 Ng6 10. d5 Nce7 11. Nc3 Nf5 12. Qc2 Nd4 13. Nxd4 cxd4 14. Ne4 Be7 15. Qd3 O-O 16. e3 Qc7 17. exd4 Rad8 18. dxe5 dxe5 19. Kh1 Bd6 20. Rfc1 Ne7 21. c5 bxc5 22. Nxc5 Bxc5 23. Bd4 e4 24. Bxc5 exd3 25. Bxe7 Qxe7 26. Rc3 Qe2 27. Rf1 d2 28. Re3 Qxf1+ 29. Bxf1 d1=Q 30. Re2
        \vspace{1mm}
        \hrule
        \vspace{1mm}
        \textbf{Answer:} \textit{Qxf1\#}\\
        \textbf{Needed Skill:} \textit{Logical Correctness, Commonsense Understanding, Completeness}\\
        \textbf{Difficulty:} \textit{Expert Level Knowledge}
        \vspace{-5pt}
    \end{tcolorbox}
    \vspace{-4mm} 
    \begin{tcolorbox}[colback=red!5, colframe=red!60!black, title=RMoA Response (Baseline)]
    \vspace{-5pt}
        I need to analyze this position after White's 30th move (Re2).\\Let me reconstruct the final position:\\- After 29...d1=Q, Black has promoted a pawn to a Queen on d1\\- White just played 30. Re2\\Current position analysis:\\- Black has a Queen on d1 (just promoted)\\- White's King is on h1\\- White just moved the Rook to e2\\- White's Bishop is on f1\\Now I need to find Black's checkmate in one move.\\Looking at White's King on h1:\\- The King is trapped in the corner\\- Black's new Queen on d1 is very powerful\\Let me check possible checkmate moves:\\**Qxf1\#** - The Queen on d1 can capture the Bishop on f1 with check. Let me verify this is checkmate:\\- The Queen on f1 gives check to the King on h1\\- The King cannot move to g1 (controlled by Queen on f1)\\- The King cannot move to g2 (controlled by Queen on f1)  \\- The King cannot move to h2 (controlled by Queen on f1)\\- The Rook on e2 cannot interpose or capture the Queen on f1\\- No other White piece can capture the Queen or block\\Yes! This is checkmate.\\**Answer: Qxf1\#**\\(31...Qxf1\# is checkmate-in-one)
        
        \vspace{1mm}
        \hrule
        \vspace{1mm}
        \textbf{FLASK Evaluation:} 
        \textit{Logical correctness}: 5, \textit{Commonsense Understanding}: 5, \textit{Completeness}: 5
        \vspace{-5pt}
    \end{tcolorbox}
    \vspace{-4mm}

    \begin{tcolorbox}[colback=blue!5, colframe=blue!60!black, title=MoA Response (Baseline)]
    \vspace{-5pt}
        Looking at this position after 30. Re2, I need to carefully reconstruct the board state:\\**White pieces:**\\- King on h1\\- Rook on e2\\- Bishop on f1\\- Pawns on a2, b3, f2, g3, h2\\**Black pieces:**\\- King on g8\\- Queen on d1 (just promoted)\\- Rooks on d8 and f8\\- Bishop on b7\\- Pawns on a7, f6, g7, h7\\It's Black to move, and I need to find checkmate in one.\\Let me examine Black's checking moves with the queen on d1:\\1. **Qxf1+** - This captures the bishop with check, but after Kg2, the king escapes.\\2. **Qd4+** - Check, but the king can move to g1 or g2.\\3. **Qg1+** - This is checkmate! Here's why:\\   - The queen on g1 gives check to the king on h1\\   - The king cannot capture on g1 (no piece can reach it)\\   - The king cannot move to g1 (occupied by the queen)\\   - The king cannot move to g2 (controlled by the queen on g1)\\   - The rook on e2 cannot interpose\\   - The bishop on f1 cannot block or capture the queen\\   - No White piece can capture the queen on g1\\**Answer: Qg1\#**\\(In full notation: 30...Qd1-g1\#)
        
        \vspace{1mm}
        \hrule
        \vspace{1mm}
        \textbf{FLASK Evaluation:} 
        \textit{Logical correctness}: 1, \textit{Commonsense Understanding}: 2, \textit{Completeness}: 3
        \vspace{-5pt}
    \end{tcolorbox}
    \vspace{-4mm}
    \begin{tcolorbox}[colback=green!5, colframe=green!50!black, title=Attention-MoA (Ours)]
    \vspace{-5pt}
        Looking at this position after 30. Re2, I need to analyze Black's pieces and White's king position.\\**Current Position:**\\- White King: h1\\- White Rook: e2\\- White Bishop: f1\\- Black Queen: d1 (just promoted)\\- Black King: g8\\Let me examine possible checking moves for Black:\\**Option 1: Qxf1+**\\- Queen captures the bishop on f1, giving check\\- Can White escape or defend?\\  - Kg2 or Kg1: Both squares are controlled by the Queen on f1\\  - Kh2: Still in check from the Queen\\  - Rxf1: The Rook on e2 cannot capture on f1 (no diagonal/straight line)\\  - No piece can block the check\\- This is checkmate! \ding{51}\\**Option 2: Qg1+**\\- This would also be checkmate IF the Queen could reach g1\\- However, the Bishop on f1 blocks the Queen's path from d1 to g1\\- This move is illegal \ding{55}\\**Verification of Qxf1\#:**\\1. The Queen captures the Bishop, delivering check to the King on h1\\2. The King has no escape squares (all controlled by the Queen)\\3. No White piece can capture the Queen or block\\4. This is mate in one move\\**Answer: Qxf1\#**
        
        \vspace{1mm}
        \hrule
        \vspace{1mm}
        \textbf{FLASK Evaluation:} 
        \textit{Logical correctness}: 5, \textit{Commonsense Understanding}: 5, \textit{Completeness}: 5
        \vspace{-5pt}
    \end{tcolorbox}
\vspace{-10pt}
    \caption{A case study comparison on the FLASK dataset involving a complex chess problem. }
    \label{fig:case_study_flask}
\end{figure*}

\vspace{-15pt}
\begin{figure*}[t]
    \centering
    \tcbset{
        width=\linewidth,      
        boxrule=0.8pt,         
        arc=2mm,               
        left=2mm, right=2mm, top=2mm, bottom=2mm, 
        fonttitle=\bfseries\small, 
        fontupper=\tiny       
    }
\vspace{-15pt}
    \begin{tcolorbox}[colback=gray!5, colframe=black!60, title=User Query]
        \vspace{-5pt}
According to Metz, what is wrong with consequentialist arguments against capital punishment based on African values?
        \vspace{1mm}
        \hrule
        \vspace{1mm}
        \textbf{Answer:} \textit{Let's think step by step. We refer to Wikipedia articles on moral disputes for help. Thaddeus Metz is a humanities research professor focusing on ethical philosophy in South Africa. Metz has written the death penalty is unjustified as it treats individuals as incapable of communal relations. It is unclear that capital punishment is to the benefit of, or a deterrent of harm to the community.}\\
        \textbf{Needed Skill:} \textit{Harmlessness, Completeness, Factuality}\\
        \textbf{Difficulty:} \textit{Expert Level Knowledge}
        \vspace{-5pt}
    \end{tcolorbox}
    \vspace{-4mm} 
    \begin{tcolorbox}[colback=red!5, colframe=red!60!black, title=RMoA (Baseline)]
    \vspace{-5pt}
        \# Metz's Critique of Consequentialist Arguments Against Capital PunishmentAccording to Thaddeus Metz, the fundamental problem with consequentialist arguments against capital punishment-even when they invoke African values-is that they **misrepresent the nature of African ethics itself**.\#\# The Core Problem: Wrong Type of Moral ReasoningMetz argues that consequentialist arguments get the **structure** of African moral reasoning wrong, even if they reach the right conclusion (opposing capital punishment).\#\#\# What a Consequentialist Argument Looks Like:A consequentialist approach based on African values might argue:-1. Actions are right if they produce good consequences-2. Good consequences = maximizing communal harmony and solidarity (ubuntu)-3. Capital punishment undermines communal harmony (creates division, prevents reconciliation, etc.)-4. Therefore, capital punishment is wrong\#\#\# Why Metz Rejects This Framework:\#\# 1. **African Ethics is Fundamentally Relational, Not Consequentialist**Metz contends that African ethics (particularly ubuntu-based ethics) is about **what constitutes harmonious relationships**, not what produces them as outcomes:- **Consequentialism asks**: "What action will *produce* the most harmony?"-- **African relationalism asks**: "What action *embodies* or *expresses* harmonious community?"The wrongness of capital punishment isn't that it has bad consequences-it's that **the act itself is inherently anti-communal**. Killing someone is the ultimate denial of their personhood and relationship to the community.\#\# 2. **Intrinsic vs. Extrinsic Wrongness**- **Consequentialism**: The wrongness is *extrinsic*-it depends on future states of affairs-- **Metz's African ethics**: The wrongness is *intrinsic*-killing violates the fundamental duty to recognize and honor shared humanityCapital punishment is wrong because it **treats a person as disposable**, severing them permanently from the community-not primarily because of its downstream effects.\#\# 3. **Misses the Deontological/Virtue-Based Character**African ethics, according to Metz, is better understood as:-- **Duty-based**: We have obligations to relate to others in certain ways (with respect, solidarity, care)-- **Virtue-based**: We should cultivate dispositions like compassion and communal identificationCapital punishment violates these duties and virtues directly, regardless of consequences.\#\# 4. **The Reconciliation Problem**Consequentialist arguments might say: "Capital punishment prevents reconciliation, which is bad."But Metz's point is deeper: **The very act of execution is itself a failure to honor the relationship**. It's not just that it prevents future reconciliation-it's that it represents the most extreme form of relational severance possible.\#\# The Correct African Argument (According to Metz)Instead of consequentialist reasoning, an authentic African argument against capital punishment would be:1. Human dignity consists in our capacity for communal relationships (ubuntu)-2. We have a duty to honor this dignity in all persons, even wrongdoers-3. Capital punishment categorically denies the offender's capacity for relationship and community membership-4. Therefore, capital punishment is intrinsically wrong-it violates our fundamental relational duties---**In summary**: Metz's critique is that consequentialist arguments, even when they use African vocabulary (like "ubuntu" or "harmony"), impose a Western utilitarian framework that distorts the essentially relational and non-consequentialist character of African moral thought. The problem isn't the conclusion but the **philosophical method** used to reach it.
        
        \vspace{1mm}
        \hrule
        \vspace{1mm}
        \textbf{FLASK Evaluation:} 
        \textit{Harmlessness}: 4, \textit{Completeness}: 4, \textit{Factuality}: 4
        \vspace{-5pt}
    \end{tcolorbox}
    \vspace{-4mm}

    \begin{tcolorbox}[colback=blue!5, colframe=blue!60!black, title=MoA (Baseline)]
    \vspace{-5pt}
\# Metz's Critique of Consequentialist Arguments Against Capital Punishment\\According to Thaddeus Metz, consequentialist arguments against capital punishment that appeal to African values suffer from a fundamental weakness:\\\#\# The Core Problem: Empirical Contingency\\**Consequentialist arguments are unstable because they depend entirely on empirical outcomes.** The main issues are:\\1. **The argument could reverse** - If evidence showed that capital punishment actually *does* promote social harmony, deter crime, or produce net positive communal outcomes, then the consequentialist would be logically committed to *supporting* it\\2. **Misses the fundamental moral wrong** - These arguments suggest capital punishment is wrong only because of its *effects* (e.g., it harms community cohesion, creates cycles of violence), not because of something intrinsically wrong with the practice itself\\3. **Not distinctively African** - The consequentialist framework doesn't capture what's unique about African moral thought\\\#\# Metz's Alternative: A Relational/Dignity-Based Argument\\Metz proposes a **non-consequentialist approach** grounded in African communitarian ethics, particularly *ubuntu*:\\- Capital punishment is wrong because it **violates human dignity** and the offender's capacity for communal relationships- It treats the person as **irredeemable** and permanently excludes them from the moral community- It fails to honor the **relational nature of personhood** central to African thought- This makes it wrong **in principle**, regardless of consequences\\\#\# Why This Matters\\Metz's approach provides a more **stable and principled** objection - capital punishment would be wrong even if it could be empirically demonstrated to have positive social effects, because it fundamentally contradicts the African ethical ideal of maintaining and restoring communal relationships.
        \vspace{1mm}
        \hrule
        \vspace{1mm}
        \textbf{FLASK Evaluation:} 
        \textit{Harmlessness}: 4, \textit{Completeness}: 4, \textit{Factuality}: 4
        \vspace{-5pt}
    \end{tcolorbox}
    \vspace{-4mm}
    \begin{tcolorbox}[colback=green!5, colframe=green!50!black, title=Attention-MoA (Ours)]
    \vspace{-5pt}
        \# Metz's Critique of Consequentialist Arguments Against Capital Punishment\\According to philosopher **Thaddeus Metz**, consequentialist arguments against capital punishment\\even when they claim to be based on African values\\suffer from two fundamental flaws:\\\#\# 1. **Empirical Contingency (The "Fragility Problem")**\\Consequentialist arguments typically claim capital punishment is wrong because it:\\- Fails to deter crime effectively\\- Costs more than life imprisonment\\- Risks executing innocent people\\- Perpetuates cycles of violence\\**Metz's objection**: These arguments make the wrongness of capital punishment **contingent on changeable empirical facts**. If evidence emerged showing that executions *did* deter crime effectively or *were* cheaper, a consistent consequentialist would be logically forced to *support* capital punishment. This reveals a philosophically weak foundation\\the opposition rests on unstable ground rather than firm moral principle.\\\#\# 2. **Philosophical Misalignment (The "Wrong Framework Problem")**\\More fundamentally, Metz argues that consequentialist reasoning **misrepresents the distinctive character of African ethics**:\\\#\#\# African Ethics is Relational, Not Consequentialist\\- **Core principle**: African moral philosophy (exemplified by *ubuntu*"I am because we are") is fundamentally about **communal relationships, shared identity, and intrinsic human dignity**\\- **Consequentialism focuses on**: Maximizing good outcomes or states of affairs\\- **The mismatch**: Consequentialist arguments locate wrongness in *bad results*, while authentic African ethics locates wrongness in acts that are **inherently discordant or anti-relational**\\\#\#\# The Intrinsic Wrong of Capital Punishment\\From an *ubuntu* perspective, capital punishment is wrong because it:\\1. **Permanently severs communal relationships** \\ the ultimate act of exclusion\\2. **Denies the offender's shared humanity** and capacity for moral transformation\\3. **Forecloses reconciliation and restoration** \\ central goals in African justice traditions\\4. **Violates the communal nature of personhood** by treating someone as disposable\\This wrongness exists **regardless of consequences**\\it's about the inherent nature of the act itself.\\\#\# Metz's Alternative\\Metz advocates for arguments grounded in **African relational ethics** that:\\- Emphasize **restoration over retribution**\\- Focus on the **intrinsic wrongness** of destroying relationship and dignity\\- Provide a **principled, non-contingent** objection to capital punishment\\- Authentically represent African philosophical traditions rather than importing Western utilitarian frameworks\\---\\**In summary**: Consequentialist arguments are both philosophically weak (contingent on empirical facts) and culturally inauthentic (imposing Western frameworks on African values). A genuine African objection would focus on how capital punishment inherently violates the relational dignity central to *ubuntu*, making it wrong in principle, not merely in practice.
        
        \vspace{1mm}
        \hrule
        \vspace{1mm}
        \textbf{FLASK Evaluation:} 
        \textit{Harmlessness}: 5, \textit{Completeness}: 5, \textit{Factuality}: 5
        \vspace{-5pt}
    \end{tcolorbox}
\vspace{-10pt}
    \caption{Qualitative comparison on the FLASK benchmark within the Humanities domain.}
    \label{fig:case_study_human}
\end{figure*}

%% file: iclr2026_conference.bib
@article{yang2025qwen3,
  title={Qwen3 technical report},
  author={Yang, An and Li, Anfeng and Yang, Baosong and Zhang, Beichen and Hui, Binyuan and Zheng, Bo and Yu, Bowen and Gao, Chang and Huang, Chengen and Lv, Chenxu and others},
  journal={arXiv preprint arXiv:2505.09388},
  year={2025}
}


%% file: ijcai26.bib
@article{
      li2024more,
      title={More agents is all you need},
      author={Li, Junyou and Zhang, Qin and Yu, Yangbin and Fu, Qiang and Ye, Deheng},
      journal={Transactions on Machine Learning Research},
      year={2024},
      url={https://openreview.net/forum?id=bgzUSZ8aeg},
      note={}
}

@article{yang2025multillmcollaborativesearchcomplex,
  title={Multi-llm collaborative search for complex problem solving},
  author={Yang, Sen and Li, Yafu and Lam, Wai and Cheng, Yu},
  journal={arXiv preprint arXiv:2502.18873},
  year={2025}
}

@inproceedings{liyun2024improving,
      author = {Du, Yilun and Li, Shuang and Torralba, Antonio and Tenenbaum, Joshua B. and Mordatch, Igor},
      title = {Improving factuality and reasoning in language models through multiagent debate},
      year = {2024},
      publisher = {JMLR.org},
      booktitle = {Proceedings of the 41st International Conference on Machine Learning},
      articleno = {467},
      numpages = {31},
      location = {Vienna, Austria},
      series = {ICML'24}
}

@inproceedings{liang-etal-2024-encouraging,
    title = "Encouraging Divergent Thinking in Large Language Models through Multi-Agent Debate",
    author = "Liang, Tian  and
      He, Zhiwei  and
      Jiao, Wenxiang  and
      Wang, Xing  and
      Wang, Yan  and
      Wang, Rui  and
      Yang, Yujiu  and
      Shi, Shuming  and
      Tu, Zhaopeng",
    editor = "Al-Onaizan, Yaser  and
      Bansal, Mohit  and
      Chen, Yun-Nung",
    booktitle = "Proceedings of the 2024 Conference on Empirical Methods in Natural Language Processing",
    month = nov,
    year = "2024",
    address = "Miami, Florida, USA",
    publisher = "Association for Computational Linguistics",
    url = "https://aclanthology.org/2024.emnlp-main.992/",
    doi = "10.18653/v1/2024.emnlp-main.992",
    pages = "17889--17904"
}

@inproceedings{chen-etal-2024-reconcile,
    title = "{R}e{C}oncile: Round-Table Conference Improves Reasoning via Consensus among Diverse {LLM}s",
    author = "Chen, Justin  and
      Saha, Swarnadeep  and
      Bansal, Mohit",
    editor = "Ku, Lun-Wei  and
      Martins, Andre  and
      Srikumar, Vivek",
    booktitle = "Proceedings of the 62nd Annual Meeting of the Association for Computational Linguistics (Volume 1: Long Papers)",
    month = aug,
    year = "2024",
    address = "Bangkok, Thailand",
    publisher = "Association for Computational Linguistics",
    url = "https://aclanthology.org/2024.acl-long.381/",
    doi = "10.18653/v1/2024.acl-long.381",
    pages = "7066--7085"
}

@inproceedings{Akbir2024Debating,
author = {Khan, Akbir and Hughes, John and Valentine, Dan and Ruis, Laura and Sachan, Kshitij and Radhakrishnan, Ansh and Grefenstette, Edward and Bowman, Samuel R. and Rockt\"{a}schel, Tim and Perez, Ethan},
title = {Debating with more persuasive LLMs leads to more truthful answers},
year = {2024},
publisher = {JMLR.org},
booktitle = {Proceedings of the 41st International Conference on Machine Learning},
articleno = {950},
numpages = {72},
location = {Vienna, Austria},
series = {ICML'24}
}

@article{nair2025tournament,
  title={Tournament of Prompts: Evolving LLM Instructions Through Structured Debates and Elo Ratings},
  author={Nair, Anirudh and Banerjee, Adi and Mombaerts, Laurent and Hagen, Matthew and Borogovac, Tarik},
  journal={arXiv preprint arXiv:2506.00178},
  year={2025}
}

@inproceedings{wang2024mixtureofagentsenhanceslargelanguage,
  title={Mixture-of-Agents Enhances Large Language Model Capabilities},
    year={2025},
  author={Wang, Junlin and Jue, WANG and Athiwaratkun, Ben and Zhang, Ce and Zou, James},
  booktitle={The Thirteenth International Conference on Learning Representations}
}

@article{li2025rethinking,
  title={Rethinking Mixture-of-Agents: Is Mixing Different Large Language Models Beneficial?},
  author={Li, Wenzhe and Lin, Yong and Xia, Mengzhou and Jin, Chi},
  journal={arXiv preprint arXiv:2502.00674},
  year={2025}
}

@inproceedings{xie-etal-2025-rmoa,
    title = "{RM}o{A}: Optimizing Mixture-of-Agents through Diversity Maximization and Residual Compensation",
    author = "Xie, Zhentao  and
      Han, Chengcheng  and
      Shi, Jinxin  and
      Cui, Wenjun  and
      Zhao, Xin  and
      Wu, Xingjiao  and
      Zhao, Jiabao",
    editor = "Che, Wanxiang  and
      Nabende, Joyce  and
      Shutova, Ekaterina  and
      Pilehvar, Mohammad Taher",
    booktitle = "Findings of the Association for Computational Linguistics: ACL 2025",
    month = jul,
    year = "2025",
    address = "Vienna, Austria",
    publisher = "Association for Computational Linguistics",
    url = "https://aclanthology.org/2025.findings-acl.342/",
    doi = "10.18653/v1/2025.findings-acl.342",
    pages = "6575--6602",
    ISBN = "979-8-89176-256-5"
}

@inproceedings{li2025smoa,
  title={SMoA: Improving Multi-agent Large Language Models with S parse M ixture-o f-A gents},
  author={Li, Dawei and Tan, Zhen and Qian, Peijia and Li, Yifan and Chaudhary, Kumar and Hu, Lijie and Shen, Jiayi},
  booktitle={Pacific-Asia Conference on Knowledge Discovery and Data Mining},
  pages={54--65},
  year={2025},
  organization={Springer}
}

@article{gao2025moma,
  title={MoMA: a mixture-of-multimodal-agents architecture for enhancing clinical prediction modelling},
  author={Gao, Jifan and Rahman, Mahmudur and Caskey, John and Oguss, Madeline and O’Rourke, Ann and Brown, Randall and Stey, Anne and Mayampurath, Anoop and Churpek, Matthew M and Chen, Guanhua and others},
  journal={npj Digital Medicine},
  year={2025},
  publisher={Nature Publishing Group UK London}
}

@inproceedings{jang2025yalenlp,
  title={Yalenlp@ peranssumm 2025: Multi-perspective integration via mixture-of-agents for enhanced healthcare qa summarization},
  author={Jang, Dongsuk and Li, Haoxin and Cohan, Arman},
  booktitle={Proceedings of the Second Workshop on Patient-Oriented Language Processing (CL4Health)},
  pages={415--427},
  year={2025}
}

@article{sharma2024patched,
  title={Patched MOA: optimizing inference for diverse software development tasks},
  author={Sharma, Asankhaya},
  journal={arXiv preprint arXiv:2407.18521},
  year={2024}
}

@article{ashiga2025industrial,
  title={Industrial llm-based code optimization under regulation: A mixture-of-agents approach},
  author={Ashiga, Mari and Voskanyan, Vardan and Dinmohammadi, Fateme and Gong, Jingzhi and Brookes, Paul and Truscott, Matthew and Giavrimis, Rafail and Basios, Mike and Kanthan, Leslie and Jie, Wei},
  journal={arXiv preprint arXiv:2508.03329},
  year={2025}
}

@article{mitra2024distributed,
  title={Distributed mixture-of-agents for edge inference with large language models},
  author={Mitra, Purbesh and Kaswan, Priyanka and Ulukus, Sennur},
  journal={arXiv preprint arXiv:2412.21200},
  year={2024}
}

@article{chen2025tumix,
  title={TUMIX: Multi-Agent Test-Time Scaling with Tool-Use Mixture},
  author={Chen, Yongchao and Chen, Jiefeng and Meng, Rui and Yin, Ji and Li, Na and Fan, Chuchu and Wang, Chi and Pfister, Tomas and Yoon, Jinsung},
  journal={arXiv preprint arXiv:2510.01279},
  year={2025}
}

@inproceedings{chakraborty2025collab,
  title={Collab: Controlled Decoding using Mixture of Agents for LLM Alignment},
  author={Chakraborty, Souradip and Bhatt, Sujay and Sehwag, Udari Madhushani and Ghosal, Soumya Suvra and Qiu, Jiahao and Wang, Mengdi and Manocha, Dinesh and Huang, Furong and Koppel, Alec and Ganesh, Sumitra},
  year={2025},
  booktitle={The Thirteenth International Conference on Learning Representations}
}

@article{ping2025verimoa,
  title={VeriMoA: A Mixture-of-Agents Framework for Spec-to-HDL Generation},
  author={Ping, Heng and Bhattacharjee, Arijit and Zhang, Peiyu and Li, Shixuan and Yang, Wei and Cheng, Anzhe and Zhang, Xiaole and Thomason, Jesse and Jannesari, Ali and Ahmed, Nesreen and others},
  journal={arXiv preprint arXiv:2510.27617},
  year={2025}
}

@article{tang2025open,
  title={Open-Source LLMs Collaboration Beats Closed-Source LLMs: A Scalable Multi-Agent System},
  author={Tang, Shengji and Cao, Jianjian and Lin, Weihao and Hong, Jiale and Zhang, Bo and Hu, Shuyue and Bai, Lei and Chen, Tao and Ouyang, Wanli and Ye, Peng},
  journal={arXiv preprint arXiv:2507.14200},
  year={2025}
}

@article{chen2025symbolic,
  title={Symbolic mixture-of-experts: Adaptive skill-based routing for heterogeneous reasoning},
  author={Chen, Justin Chih-Yao and Yun, Sukwon and Stengel-Eskin, Elias and Chen, Tianlong and Bansal, Mohit},
  journal={arXiv preprint arXiv:2503.05641},
  year={2025}
}

@article{chen2024more,
  title={Are more llm calls all you need? towards the scaling properties of compound ai systems},
  author={Chen, Lingjiao and Davis, Jared and Hanin, Boris and Bailis, Peter and Stoica, Ion and Zaharia, Matei and Zou, James},
  journal={Advances in Neural Information Processing Systems},
  volume={37},
  pages={45767--45790},
  year={2024}
}

@article{li2025llms,
  title={Llms can generate a better answer by aggregating their own responses},
  author={Li, Zichong and Feng, Xinyu and Cai, Yuheng and Zhang, Zixuan and Liu, Tianyi and Liang, Chen and Chen, Weizhu and Wang, Haoyu and Zhao, Tuo},
  journal={arXiv preprint arXiv:2503.04104},
  year={2025}
}

@article{ji2025survey,
  title={A Survey of Test-Time Compute: From Intuitive Inference to Deliberate Reasoning},
  author={Ji, Yixin and Li, Juntao and Xiang, Yang and Ye, Hai and Wu, Kaixin and Yao, Kai and Xu, Jia and Mo, Linjian and Zhang, Min},
  journal={arXiv preprint arXiv:2501.02497},
  year={2025}
}

@article{tran2025multi,
  title={Multi-agent collaboration mechanisms: A survey of llms},
  author={Tran, Khanh-Tung and Dao, Dung and Nguyen, Minh-Duong and Pham, Quoc-Viet and O'Sullivan, Barry and Nguyen, Hoang D},
  journal={arXiv preprint arXiv:2501.06322},
  year={2025}
}

@inproceedings{Guo2024multiagentsurveyijcai,
author = {Guo, Taicheng and Chen, Xiuying and Wang, Yaqi and Chang, Ruidi and Pei, Shichao and Chawla, Nitesh V. and Wiest, Olaf and Zhang, Xiangliang},
title = {Large language model based multi-agents: a survey of progress and challenges},
year = {2024},
isbn = {978-1-956792-04-1},
url = {https://doi.org/10.24963/ijcai.2024/890},
doi = {10.24963/ijcai.2024/890},
booktitle = {Proceedings of the Thirty-Third International Joint Conference on Artificial Intelligence},
articleno = {890},
numpages = {10},
location = {Jeju, Korea},
series = {IJCAI '24}
}

@inproceedings{Wei2025ijcaiContextAware,
author = {Wei, Zheng and Li, Mingchen and Zhang, Pu and Liu, Xinyu and Qu, Huamin and Hui, Pan},
title = {ContextAware: a multi-agent framework for detecting harmful image-based comments on social media},
year = {2025},
isbn = {978-1-956792-06-5},
url = {https://doi.org/10.24963/ijcai.2025/1103},
doi = {10.24963/ijcai.2025/1103},
booktitle = {Proceedings of the Thirty-Fourth International Joint Conference on Artificial Intelligence},
articleno = {1103},
numpages = {9},
location = {Montreal, Canada},
series = {IJCAI '25}
}

@article{rastogi2025magistral,
  title={Magistral},
  author={Rastogi, Abhinav and Jiang, Albert Q and Lo, Andy and Berrada, Gabrielle and Lample, Guillaume and Rute, Jason and Barmentlo, Joep and Yadav, Karmesh and Khandelwal, Kartik and Chandu, Khyathi Raghavi and others},
  journal={arXiv preprint arXiv:2506.10910},
  year={2025}
}

@article{team2025gemma,
  title={Gemma 3 technical report},
  author={Team, Gemma and Kamath, Aishwarya and Ferret, Johan and Pathak, Shreya and Vieillard, Nino and Merhej, Ramona and Perrin, Sarah and Matejovicova, Tatiana and Ram{\'e}, Alexandre and Rivi{\`e}re, Morgane and others},
  journal={arXiv preprint arXiv:2503.19786},
  year={2025}
}

@article{agarwal2025gpt,
  title={gpt-oss-120b \& gpt-oss-20b model card},
  author={Agarwal, Sandhini and Ahmad, Lama and Ai, Jason and Altman, Sam and Applebaum, Andy and Arbus, Edwin and Arora, Rahul K and Bai, Yu and Baker, Bowen and Bao, Haiming and others},
  journal={arXiv preprint arXiv:2508.10925},
  year={2025}
}

@misc{meta2025llama4,
  author = {Meta},
  year = {2025},
  title = {The Llama 4 herd: The beginning of a new era of natively multimodal intelligence},
  howpublished = {\url{https://ai.meta.com/blog/llama-4-multimodal-intelligence/}},
}

@techreport{anthropic2025claude45,
  title       = {Claude 4.5 Sonnet System Card},
  author      = {Anthropic},
  institution = {Anthropic PBC},
  year        = {2025},
  url         = {https://assets.anthropic.com/m/12f214efcc2f457a/original/Claude-Sonnet-4-5-System-Card.pdf},
  note        = {Technical Report}
}

@article{comanici2025gemini,
  title={Gemini 2.5: Pushing the frontier with advanced reasoning, multimodality, long context, and next generation agentic capabilities},
  author={Comanici, Gheorghe and Bieber, Eric and Schaekermann, Mike and Pasupat, Ice and Sachdeva, Noveen and Dhillon, Inderjit and Blistein, Marcel and Ram, Ori and Zhang, Dan and Rosen, Evan and others},
  journal={arXiv preprint arXiv:2507.06261},
  year={2025}
}

@article{liu2024deepseek,
  title={Deepseek-v3 technical report},
  author={Liu, Aixin and Feng, Bei and Xue, Bing and Wang, Bingxuan and Wu, Bochao and Lu, Chengda and Zhao, Chenggang and Deng, Chengqi and Zhang, Chenyu and Ruan, Chong and others},
  journal={arXiv preprint arXiv:2412.19437},
  year={2024}
}

@misc{openai2025gpt41,
  author = {{OpenAI}},
  year = {2025},
  title = {Introducing GPT-4.1},
  howpublished = {\url{https://openai.com/index/gpt-4-1/}},
}

@misc{qwen3max,
    title = {Qwen3-Max: Just Scale it},
    author = {Qwen Team},
    month = {September},
    year = {2025}
}

@article{dubois2024length,
  title={Length-Controlled AlpacaEval: A Simple Way to Debias Automatic Evaluators},
  author={Dubois, Yann and Galambosi, Bal{\'a}zs and Liang, Percy and Hashimoto, Tatsunori B},
  journal={arXiv preprint arXiv:2404.04475},
  year={2024}
}

@article{zheng2023judging,
  title={Judging llm-as-a-judge with mt-bench and chatbot arena},
  author={Zheng, Lianmin and Chiang, Wei-Lin and Sheng, Ying and Zhuang, Siyuan and Wu, Zhanghao and Zhuang, Yonghao and Lin, Zi and Li, Zhuohan and Li, Dacheng and Xing, Eric and others},
  journal={Advances in neural information processing systems},
  volume={36},
  pages={46595--46623},
  year={2023}
}

@inproceedings{ye2024flask,
  title={FLASK: Fine-grained Language Model Evaluation based on Alignment Skill Sets},
  author={Ye, Seonghyeon and Kim, Doyoung and Hwang, Hyeongbin and Kim, Seungone and Jo, Yongrae and Thorne, James and Kim, Juho and Seo, Minjoon},
  booktitle={ICLR 2024},
  year={2024},
  organization={International Conference on Learning Representations (ICLR)}
}

@inproceedings{liu2025survey,
  title={A survey on the feedback mechanism of LLM-based AI agents},
  author={Liu, Zhipeng and Bai, Xuefeng and Chen, Kehai and Chen, Xinyang and Li, Xiucheng and Xiang, Yang and Liu, Jin and Li, Hong-Dong and Wang, Yaowei and Nie, Liqiang and others},
  booktitle={Proceedings of the Thirty-Fourth International Joint Conference on Artificial Intelligence},
  pages={10582--10592},
  year={2025},
  organization={International Joint Conferences on Artificial Intelligence}
}

@inproceedings{xu2025generative,
  title={Generative agents for multimodal controversy detection},
  author={Xu, Tianjiao and Gao, Jinfei and Kong, Keyi and Yin, Jianhua and Gan, Tian and Nie, Liqiang},
  booktitle={Proceedings of the Thirty-Fourth International Joint Conference on Artificial Intelligence},
  pages={9963--9971},
  year={2025}
}
